\documentclass[lettersize,journal]{IEEEtran}
\usepackage{amsmath,amsfonts}
\usepackage{algorithmic}
\usepackage{algorithm}
\usepackage{array}
\usepackage[caption=false,font=normalsize,labelfont=sf,textfont=sf]{subfig}
\usepackage{textcomp}
\usepackage{stfloats}
\usepackage{url}
\usepackage{verbatim}
\usepackage{graphicx}
\usepackage{cite}
\usepackage{textcomp}
\usepackage{stfloats}
\usepackage{url}
\usepackage{verbatim}
\usepackage{graphicx}
\usepackage{cite}
\usepackage{color}
\usepackage{xcolor}
\usepackage{booktabs}
\usepackage{multirow}
\usepackage{pifont}
\usepackage{makecell}
\usepackage{algorithmic}
\usepackage{graphicx}
\usepackage{textcomp}
\usepackage{xcolor}
\usepackage{url}
\usepackage{cite}
\usepackage{amsmath,amssymb,amsfonts}
\usepackage{amsmath,amsfonts}
\usepackage{algorithmic}
\usepackage{algorithm}
\usepackage{array}
\usepackage{tikz,xcolor}
\usepackage[implicit=false]{hyperref}
\hypersetup{hidelinks,
	colorlinks=true,
	allcolors=black,
	pdfstartview=Fit,
	breaklinks=true}
	
\definecolor{lime}{HTML}{A6CE39}
\DeclareRobustCommand{\orcidicon}{
\begin{tikzpicture}
\draw[lime, fill=lime] (0,0)
circle[radius=0.13]
node[white]{{\fontfamily{qag}\selectfont \tiny \.{I}D}};
\end{tikzpicture}
\hspace{-2mm}
}
\foreach \x in {A, ..., Z}{%
\expandafter\xdef\csname orcid\x\endcsname{\noexpand\href{https://orcid.org/\csname orcidauthor\x\endcsname}{\noexpand\orcidicon}}
}

\begin{document}

\title{Topology-Driven Attribute Recovery for Attribute Missing Graph Learning in Social Internet of Things}

\author{
Mengran~Li\hspace{-2mm}\orcidA{}\hspace{-1mm},
Junzhou~Chen\hspace{-2mm}\orcidG{}\hspace{-1mm},
Chenyun Yu\hspace{-2mm}\orcidF{}\hspace{-1mm},
Guanying~Jiang\hspace{-2mm}\orcidD{}\hspace{-1mm},
Ronghui~Zhang\hspace{-2mm}\orcidB{}\hspace{-1mm},
Yanming Shen\hspace{-2mm}\orcidC{}\hspace{-1mm},
and Houbing Herbert Song\hspace{-2mm}\orcidE{}\hspace{-1mm},~\IEEEmembership{Fellow,~IEEE}

\IEEEcompsocitemizethanks{

\IEEEcompsocthanksitem This work has been submitted to the lEEE for possible publicationCopyright may be transferred without notice, after which this version mayno longer be accessible.

\IEEEcompsocthanksitem This project is jointly supported by the Shenzhen Fundamental Research Program (No. JCYJ20240813151129038), the National Natural Science Foundation of China (Nos. 52172350, 51775565), the Guangdong Basic and Applied Research Foundation (No. 2022B1515120072), the Guangzhou Science and Technology Plan Project (No. 2024B01W0079), the Nansha Key R\&D Program (No. 2022ZD014). \textit{(Corresponding author: Ronghui Zhang.)}
\IEEEcompsocthanksitem  Mengran Li, Junzhou Chen, Chenyun Yun, and Ronghui~Zhang are with School of Intelligent Systems Engineering, Shenzhen Campus of Sun Yat-sen University, Shenzhen 518107, P.R. China. (e-mail: limr39@mail2.sysu.edu.cn, chenjunzhou@mail.sysu.edu.cn, yuchy35@mail.sysu.edu.cn, zhangrh25@mail.sysu.edu.cn).
\IEEEcompsocthanksitem 
Guanying Jiang is with the Baidu Inc, Beijing 100085, P.R. China. (e-mail: guanyingjiang@gmail.com).
\IEEEcompsocthanksitem  
Yanming~Shen is with the Key Laboratory of Intelligent Control and Optimization for Industrial Equipment, Ministry of Education, the School of Electronic Information and Electrical Engineering, Dalian University of Technology, Dalian 116024, P.R. China. (e-mail: shen@dlut.edu.cn).
\IEEEcompsocthanksitem  
Hongbing Herbert Song is with the Security and Optimization for Networked Globe Laboratory (SONG Lab), Department of Information Systems, University of Maryland, Baltimore County, Baltimore, MD 21250 USA. (e-mail: h.song@ieee.org).

% Manuscript received April 19, 2023; revised August 16, 2023.
}
}



\maketitle

\begin{abstract}

With the advancement of information technology, the Social Internet of Things (SIoT) has fostered the integration of physical devices and social networks, deepening the study of complex interaction patterns. Text Attribute Graphs (TAGs) capture both topological structures and semantic attributes, enhancing the analysis of complex interactions within the SIoT. However, existing graph learning methods are typically designed for complete attributed graphs, and the common issue of missing attributes in Attribute Missing Graphs (AMGs) increases the difficulty of analysis tasks. To address this, we propose the Topology-Driven Attribute Recovery (TDAR) framework, which leverages topological data for AMG learning. TDAR introduces an improved pre-filling method for initial attribute recovery using native graph topology. Additionally, it dynamically adjusts propagation weights and incorporates homogeneity strategies within the embedding space to suit AMGs' unique topological structures, effectively reducing noise during information propagation. Extensive experiments on public datasets demonstrate that TDAR significantly outperforms state-of-the-art methods in attribute reconstruction and downstream tasks, offering a robust solution to the challenges posed by AMGs. The code is available at \url{https://github.com/limengran98/TDAR}.
\end{abstract}


\section{Introduction}
\begin{figure}[t]
\centering
\includegraphics[width=0.48\textwidth]{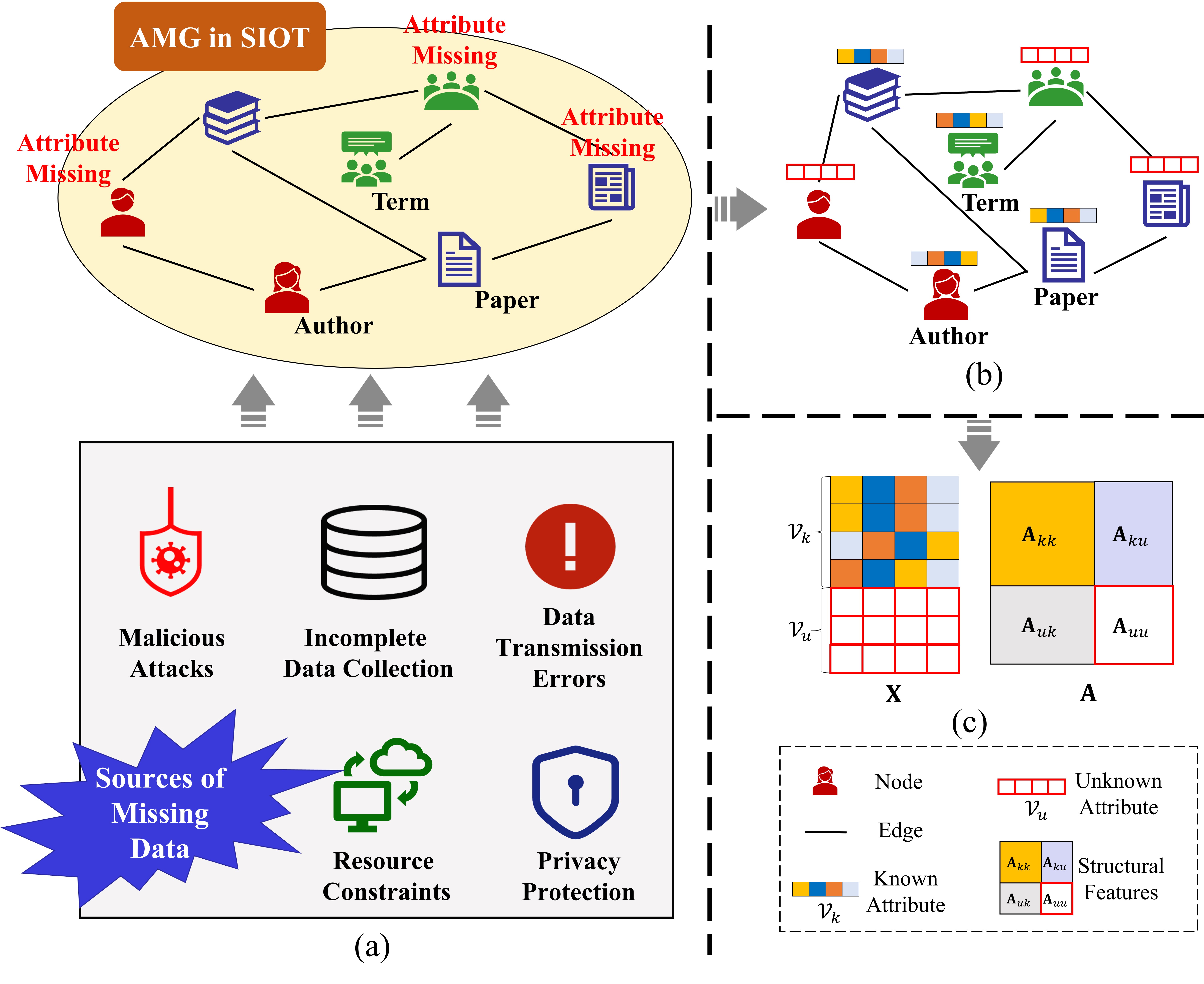}
\caption{Representation of AMG and its features in a citation network of the SIoT. (a) Illustration of data missing in SIoT, (b) Illustration of the AMG, (c) Known and unknown attributes as well as structural features of AMG.}
\label{fig0}
\end{figure}

\IEEEPARstart{W}{ith} the rapid advancement of technologies like artificial intelligence, the Internet of Things (IoT), and social media, the integration of physical and social domains has become increasingly prominent \cite{chen2021zero, li2021dual, wang2022minority}. Among them, the Social Internet of Things (SIoT) particularly focuses on the social aspects within IoT and is widely applied in various scenarios such as social networks \cite{li2022multi, li2023hypergraph}, privacy protection \cite{zuo2020privacy, tian2021achieving}, and recommendation systems \cite{wu2022eagcn, jing2023dual}. In SIoT, Text Attribute Graphs (TAGs) are widely employed to describe and analyze community relationships, as they can simultaneously capture both the topological structure of edges and the semantic attributes of nodes \cite{song2022graph, zhou2022graph, li2023self, li2024redundancy}. However, due to issues such as privacy protection and incomplete data collection, TAGs often suffer from missing attributes, known as Attribute Missing Graphs (AMGs), as shown in Figure \ref{fig0} (a). Since most graph learning methods and models are designed for complete attributed graphs, AMGs present significant challenges to traditional graph learning \cite{you2020handling, rossi2022unreasonable, guo2023fair}. As shown in Figure \ref{fig0} (b), entities in TAGs often have incomplete attribute features.

Effectively analyzing AMGs has become a pressing problem. Traditionally, reconstructing missing attributes relies on statistical and linear algebra methods, such as feature interpolation \cite{silva2015single, xia2017adjusted} and matrix completion \cite{cai2010singular, candes2012exact}. However, these methods perform poorly when handling unstructured graph data. To address these limitations, graph neural networks (GNNs) like graph convolutional networks (GCN) \cite{kipf2016semi}, graph attention networks (GAT) \cite{velivckovic2017graph}, and GraphSAGE \cite{hamilton2017inductive} have been developed to leverage graph topology for attribute propagation between nodes, but these methods are not specifically designed for AMGs and cannot effectively recover missing attributes, resulting in poor performance.

Recognizing this gap, researchers have turned to generative techniques to better handle missing data in AMGs. Methods like autoencoders (AEs) \cite{kingma2013auto}, variational graph autoencoders (VGAEs) \cite{kipf2016variational}, and generative adversarial networks (GANs) \cite{goodfellow2020generative} enhance attribute learning by embedding auxiliary tasks. For example, SAT \cite{chen2022learning, jin2022amer} combines AEs and GANs, using dual encoders to learn both attribute and structural representations of AMGs and generating missing attributes using structural representations. SVGA \cite{yoo2022accurate} employs a structured variational graph autoencoder architecture and incorporates Gaussian Markov random fields (GMRFs) \cite{lauritzen1988local} into the distribution of latent variables to accurately estimate node features. In addition, methods like CSAT \cite{li2023csat} and AmGCL \cite{zhang2023amgcl} introduce contrastive learning into AMG processing, utilizing mutual information maximization to generate rich node representations and enhance model robustness. To further refine attribute recovery, ITR \cite{tu2022initializing} initially fills missing attributes using graph structural information, followed by adaptive refinement. MATE \cite{peng2023multi} integrates multi-view information to impute missing graph attributes, enhancing the effectiveness and robustness of attribute imputation through a multi-view fusion strategy. AIAE \cite{xia2024attribute} employs a dual encoder based on knowledge distillation to more effectively integrate existing attributes and structural information during the encoding stage and enhances decoding capability with a multi-scale decoder featuring masking.

Despite the effectiveness of existing methods in certain contexts, they face notable limitations and challenges. First, existing methods often neglect the proper initialization of missing node attributes before inputting them into the network. Arbitrary or inaccurate initialization of these missing attributes can introduce significant noise, adversely affecting the representations of nodes with known attributes. Second, as illustrated in Figure \ref{fig0} (c), the topology of AMGs is more complex than that of fully attributed graphs. AMGs feature four types of node relationships based on whether nodes have known or unknown attributes: known-known ($\mathbf A_{kk}$), known-unknown ($\mathbf A_{ku}$), unknown-known ($\mathbf A_{uk}$), and unknown-unknown ($\mathbf A_{uu}$). Existing methods often treat all nodes uniformly, failing to account for the varying importance and influence of different node types, which leads to inaccuracies in recovering missing attributes. Third, during the learning process of AMGs, the embedding space faces issues of attribute dispersion between connected nodes and misleading attribute similarities between non-connected nodes. Current methods lack robust mechanisms to distinguish whether variations in node similarity of embedding space stem from the true underlying structure of the graph or are artifacts introduced by missing data.

To overcome these challenges, we propose a novel framework named Topology-Driven Attribute Recovery (TDAR), which leverages graph topology through a multitask learning mechanism to more comprehensively and accurately handle AMGs. TDAR introduces three key solutions to address the challenges of AMGs:

\textbf{Topology-Aware Attribute Propagation (TAAP):} TAAP views attribute propagation as a Dirichlet energy minimization process. By utilizing the graph's topological features, it propagates known node attributes to pre-fill those of nodes with missing attributes. TAAP employs both global propagation, ensuring globality across the graph, and a known reset mechanism, which prevents the distortion of original known attributes during the process. This approach directly tackles the challenge of initializing missing attributes, reducing noise and improving feature quality.

\textbf{Embedding Space Propagation Confidence (ESPC):} To adjust attribute recovery based on node importance, the ESPC strategy assigns dynamic attention weights to nodes according to their topological positions in the graph (known or unknown attributes). This method compensates for potential errors in information propagation, ensuring more accurate modeling of diverse topological relationships within AMGs.

\textbf{Node Homogeneity Score (NHS) and Non-Linkage Similarity Calibration (NLSC):} NHS evaluates homogeneity among neighboring nodes, ensuring similar nodes cluster together in the embedding space, promoting coherence. NLSC calibrates misleading similarities between non-connected nodes that may appear similar in the embedding space, preventing incorrect interpretations and preserving structural integrity.

In summary, TDAR not only improves the initialization of missing attributes but also intelligently models the complex topological relationships inherent in AMGs, all while effectively filtering out noise and redundancy. Our contributions include the following innovations:
\begin{itemize}
    \item We introduce the TAAP, which utilizes the topological structure of the graph to pre-fill missing node attributes. Unlike conventional methods, TAAP leverages global graph connectivity to guide context-aware pre-filling, reducing noise in generative models and ensuring higher-quality attribute recovery from the outset.
    
    \item We propose the ESPC mechanism to account for the varying significance of nodes with both known and missing attributes. ESPC dynamically adjusts node influence by assigning attention weights based on each node's topological position.
    
    \item We enhance graph embedding by aligning node similarities with their topological relationships through the NHS and NLSC. These mechanisms mitigate the impact of noisy or redundant connections, ensuring that both connected and non-connected nodes are represented accurately in the embedding space.

   \item Extensive experiments on public datasets validate the effectiveness of our framework, demonstrating superior performance in attribute reconstruction, node classification, and clustering tasks compared to existing state-of-the-art methods.
\end{itemize}
The organization of this paper is as follows: Section II provides a review of related work. Section III describes the problem and outlines the fundamental methods. Section IV presents our proposed framework. Section V examines the experimental outcomes, and Section VI summarizes the findings and suggests directions for future research.

\section{Related Work}
\subsection{Deep Graph Learning}
Deep graph learning performs well in graph community detection and analysis. GNNs \cite{gori2005new, scarselli2008graph} have unique advantages in capturing complex relationships and features in graph structures, significantly improving the performance of tasks such as community detection, node classification, and link prediction. GCN \cite{kipf2016semi} efficiently learns node representations through first-order approximate graph convolution operations and is widely used in node classification and graph embedding. GAT \cite{velivckovic2017graph} introduces an attention mechanism to dynamically weight the features of neighboring nodes according to their importance, thereby improving the flexibility and accuracy of node representation. GraphSAGE \cite{hamilton2017inductive} generates node embeddings by sampling and aggregating node neighborhood features, enabling it to handle large-scale graph data. In terms of self-supervised models, GAE \cite{kipf2016variational} applies autoencoders to graph data, learns node embeddings by encoding node features and reconstructing graph structures. VGAE \cite{kipf2016variational} combines the advantages of VAE \cite{kingma2013auto} and GAE, and can better handle the uncertainty and complex structure in graph data by learning the implicit distribution of node features through variational inference. Recent methods such as Graph Transformer \cite{shi2020masked} use the Transformer architecture to capture global and local information in graph data. GraphGPS \cite{rampavsek2022recipe} is committed to enhancing the ability of graph neural networks (GNNs) in capturing graph structure information and node positions. GraphGPS improves the performance of the model in various graph-related tasks by seamlessly combining two types of information: global position encoding and local structure encoding.

Although these methods are highly effective for general graph representation learning, they are not specifically designed to address the challenges posed by AMGs. As a result, they struggle to recover missing attributes, which is critical for accurate analysis in AMG contexts. Our TDAR framework not only builds upon the foundational concepts of methods like GCN and GAE, utilizing message passing for missing attribute reconstruction, but also introduces more advanced strategies that address the unique challenges presented by AMGs.

\subsection{Attribute Missing Graph Learning}
To address the issue of missing node attributes in graph learning, diverse deep learning strategies have been explored. Early methods focused on simple feature aggregation techniques. Simcsek \emph{et al.} \cite{csimcsek2008navigating} utilized mean pooling to aggregate neighboring node features. Huang \cite{huang2019graph} and Chen \emph{et al.} \cite{chen2019attributed} employed techniques based on attributed random walks for generating node embeddings within bipartite graphs that include node attributes. 

Generative models have been widely used for handling missing data. Yoon \emph{et al.} \cite{yoon2018gain} applied GANs \cite{goodfellow2020generative} to synthesize absent data, thereby mirroring the real data's distribution. Vincent \emph{et al.} \cite{vincent2010stacked} and Spinelli \emph{et al.} \cite{spinelli2020missing} innovatively introduced a denoising  graph auto-encoder (GAE), encoding edge-based similarity patterns to extract robust features and complete missing attribute reconstruction. Taguchi \emph{et al.} \cite{taguchi2021graph} transformed missing attributes into Gaussian mixture models, enabling the use of GCNs in networks with incomplete attributes. Chen \emph{et al.} \cite{chen2022learning}, Jin \emph{et al.} \cite{jin2022amer}, and Yoo \emph{et al.} \cite{yoo2022accurate} focused on learning and probabilistic modeling of node representations. They built on dual auto-encoder architectures and used techniques like GANs \cite{goodfellow2020generative} and GMRFs \cite{lauritzen1988local}, under the guidance of shared latent space assumptions and structural information. 

Recently, self-supervised learning has gained traction in node representation learning for AMGs. Li \emph{et al.} \cite{li2023csat} leveraged contrastive learning to identify patterns among nodes, aggregating information from diverse samples, and employing a Transformer architecture for modeling node relationships. Zhang \emph{et al.} \cite{zhang2023amgcl} introduced a novel self-supervised learning structure using contrastive learning, effectively handling missing node attributes in attribute graphs. Huo \emph{et al.} \cite{huo2023t2} effectively improved the performance of GNNs on incomplete graphs by separately designing feature-level and structure-level teacher models and avoiding interference between features and structures through a distillation process.

Several methods have employed autoencoders to address missing attributes by leveraging the graph's structure. Tu \emph{et al.} \cite{tu2022initializing} proposed an imputation initialization method to initially fill missing attributes by leveraging the graph's structural information, followed by adaptive improvement of the imputed latent variables. Peng \emph{et al.} \cite{peng2023multi} developed an attribute imputation method for the input space, incorporating parameter initialization and graph diffusion to generate multi-view information. Tenorio \emph{et al.} \cite{tenorio2024recovering} introduced a method for recovering node features in graph data by using graph autoencoders and considering local graph structure, effectively addressing the problem of completely missing node features. Xia \emph{et al.} \cite{xia2024attribute} proposed a novel attribute imputation autoencoder  that employs a dual encoder based on knowledge distillation and a multi-scale decoder with masking to improve the accuracy, robustness, and generative ability of imputation for AMGs.

Although existing methods relying on generative models have made significant progress in AMG learning, several key challenges remain unresolved, such as the pre-filling of initial attribute features, the impact of missing attribute nodes on confidence in the embedding space, and the topological guidance for attribute reconstruction. To more fully leverage the rich context provided by the graph's topological propagation and embedding space,TDAR leverages native graph topology to propagate known attributes and pre-fill missing ones, ensuring both global consistency and local refinement. It dynamically adjusts propagation weights based on node importance, capturing relationships across different graph scales. Additionally, it enhances robustness by aligning the embedding space with the graph's structure and attributes, leading to more accurate attribute reconstruction.

\subsection{Feature Propagation for Attribute Recovery}
Feature Propagation (FP), a simple and efficient method, has been applied in AMG learning, primarily relying on the principle of Dirichlet energy minimization \cite{shannon2001mathematical}. Dirichlet energy minimization ensures that attributes in the graph change smoothly across its structure. This implies that the attributes of a node are influenced by, and tend to be similar to, those of its neighboring nodes. Therefore, this method can efficiently and effectively pre-fill missing attributes.
Rossi \emph{et al.} \cite{rossi2022unreasonable} utilized FP to leverage graph structures, while Um \emph{et al.} \cite{um2023confidence} extended this method by introducing a pseudo-confidence weighted mechanism for feature propagation. 

However, traditional methods such as FP often require multiple iterations, which may lead to excessive feature smoothing, and known attribute features are not updated during iteration. Our method introduces a small global average value during the update of unknown attributes and allows known attribute features to participate in the iteration process. This reduces the number of iterations and refines feature processing, thereby enhancing efficiency and accuracy.

\subsection{Graph Learning in SIoT}
The development of the SIoT has driven researchers to explore the social relationships between IoT devices and their application scenarios \cite{malekshahi2020social}. Jung \emph{et al.} \cite{jung2018quantitative} proposed a method to quantify social strength in SIoT, enhancing system performance by analyzing the social relationships between devices. Khelloufi \emph{et al.} \cite{khelloufi2020social} designed a service recommendation system based on social relationships, focusing on service matching and personalized recommendations between devices. Guo \emph{et al.} \cite{guo2021deep} proposed a deep learning-embedded SIoT system to address ambiguity-aware social recommendations. By leveraging social relationships between devices and deep learning models, the system provides more intelligent recommendation services. Sun \emph{et al.} \cite{sun2021integrated} proposed an integrated PCA and DAEGCN model for movie recommendation, leveraging both social and physical network information in SIoT.  Yang \emph{et al.} \cite{yang2021learning} introduced a learning-driven task-optimized group search algorithm, which improves task allocation and execution efficiency in the SIoT environment by optimizing the search process between devices. Mohana \emph{et al.} \cite{mohana2023ccnsim} developed an AI-powered classification, clustering, and navigation simulator (CCNSim) for SIoT, aiming to enhance communication efficiency and navigation capabilities between devices, advancing the intelligence of SIoT systems. Jing \emph{et al.} \cite{jing2023dual} designed a dual preference perception network, targeting fashion recommendation scenarios and further enhancing the personalization level of recommendation systems in SIoT.

In the field of graph learning, researchers have been exploring how to utilize graph-structured data to improve various tasks. Wu \emph{et al.} \cite{wu2022eagcn} proposed an efficient adaptive graph convolutional network (EAGCN), which was applied to item recommendation in SIoT, effectively improving recommendation accuracy and adaptability. Wang \emph{et al.} \cite{wang2022minority} designed a minority-weighted graph neural network model that focuses on addressing the imbalance in node classification in social networks, ensuring that the information from minority nodes is better learned and captured. Additionally, Chen \emph{et al.} \cite{chen2021zero} utilized knowledge graph embeddings to achieve zero-shot text classification, particularly suited for label-free classification scenarios in social media data. Li \emph{et al.} \cite{li2021dual} further explored how to improve node classification performance in social networks through the dual mutual robust graph convolutional network in weakly supervised settings. Furthermore, Jiang \emph{et al.} \cite{jiang2024tfd} proposed a trust-based fraud detection model (TFD), which utilizes graph GCN to analyze device interactions in SIoT.

SIoT and graph learning are closely related research fields, where the social relationships and interactions between IoT devices naturally form complex graph structures. However, the issue of missing data limits the effectiveness of traditional graph learning methods when handling the complex graph structures in SIoT. To address this, we propose an effective AMG handling strategy aimed at filling in the missing attributes in SIoT, thereby further enhancing the applicability and potential of graph learning techniques in SIoT.

\section{Preliminaries}

\subsection{Problem Definition}
We define a graph $G=(V,E)$ with missing node attributes. Here, $V=V_u\cup V_k$ represents the node set, where $V_k$ and $V_u$ denote nodes with unknown and known attributes, respectively. Among a total of $N$ nodes, only $k$ nodes possess attributes, satisfying $|k|+|u|=N$. $E$ represents the set of edges, and $M$ denotes the number of edges. The node attribute feature matrix is represented as $ \mathbf X = \begin{bmatrix}\mathbf {X}_k \\\mathbf{X}_u \end{bmatrix} \in \mathbb{R}^{N \times F} $, where $F$ is the number of attribute features. Other relevant information includes the adjacency matrix $ \mathbf A = \begin{bmatrix}
\mathbf{A}_{kk} & \mathbf A_{ku} \\
\mathbf A_{uk} & \mathbf A_{uu} 
\end{bmatrix} \in \{0,1\}^{N \times N} $, the degree matrix $\mathbf{D}$, symmetrically normalized adjacency matrix $ \hat{\mathbf A} = \mathbf D^{-1/2}\mathbf{AD}^{-1/2} $, and the Laplacian matrix  $ \mathbf L = \mathbf I - \hat{\mathbf A} $. The objective of this work is to reconstruct the attribute features $\hat{\mathbf X}$ of the graph with missing attributes and apply them to downstream tasks.

\subsection{Feature Propagation}
FP \cite{rossi2022unreasonable, um2023confidence} considers attribute propagation as a process of Dirichlet energy minimization \cite{shannon2001mathematical}, enhancing feature smoothness across the topology \cite{oono2019graph}. Using the graph Laplacian matrix $\mathbf{L}$, FP can estimate the properties of unknown nodes from the known nodes. FP aims to find a smooth attribute distribution that minimizes changes on the graph, achievable by minimizing the Dirichlet energy function $\mathbb E(\mathbf{X}) = \mathbf{X}^T \mathbf{L} \mathbf{X}$. Setting its derivative to zero, we have:
\begin{equation}
\begin{gathered}
\mathbb E(\mathbf X) =\begin{bmatrix}
\mathbf{X}_k^T & \mathbf{X}_u^T
\end{bmatrix}\begin{bmatrix}
\mathbf{L}_{k k} & \mathbf{L}_{k u} \\
\mathbf{L}_{u k} & \mathbf{L}_{u u}
\end{bmatrix}\begin{bmatrix}
\mathbf{X}_k \\
\mathbf{X}_u
\end{bmatrix}\\
\frac{\partial \mathbb E (\mathbf X)}{\partial \mathbf{X}_u}=0 \\
\mathbf{L}_{u k} \mathbf{X}_k+\mathbf{L}_{u u} \mathbf{X}_u=0 \\
\mathbf{X}_u=-\mathbf{L}_{u u}^{-1} \mathbf{L}_{u k} \mathbf{X}_k.
\end{gathered}
\end{equation}

Due to the high complexity of matrix inversion, we can opt for a discrete iterative approach to solve it, assuming the boundary condition $\mathbf X_k^{(l)} = \mathbf X_k^{(0)}$, we have:
\begin{equation}\label{e5}
\mathbf{X}^{(l+1)}=\mathbf{X}^{(l)}-\begin{bmatrix}\mathbf{0} & \mathbf{0} \\ \mathbf{L}_{u k} & \mathbf{L}_{u u} \end{bmatrix} \mathbf{X}^{(l)} = \begin{bmatrix} \mathbf{I} & \mathbf{0} \\ \hat{\mathbf{A}}_{u k} & \hat{\mathbf{A}}_{u u}
\end{bmatrix}\begin{bmatrix}
\mathbf{X}_k^{(l)} \\
\mathbf{X}_u^{(l)}
\end{bmatrix}.\\
\end{equation}
The attribute aggregation process, governed by the propagation layers $l$, is iteratively optimized.

FP uses an iterative process to update the attribute vector $\mathbf{X}$, gradually approaching the optimal solution, which can be expressed as:
\begin{equation}\label{e6}
\begin{aligned}
&\tilde{\mathbf X}_u^{(l+1)} = \hat{\mathbf A}_{uk} \mathbf X_k^{(l)} + \hat{\mathbf A}_{uu} \mathbf X_u^{(l)}\\
&\tilde{\mathbf X}_k^{(l+1)} = {\mathbf X}_k^{(0)},
\end{aligned}
\end{equation}
where $\tilde{\mathbf X} =  \begin{bmatrix}\tilde{\mathbf {X}}_k \\\tilde{\mathbf{X}}_u \end{bmatrix}$ represents the refined attribute feature. Equation \eqref{e6} facilitates the transfer of attribute information from known to unknown nodes in the recovery process, re-expressed as refined attributes $\tilde{\mathbf X}$, while the attributes of the known nodes remain unaltered.

\begin{figure*}[t]
\centering
\includegraphics[width=0.95\textwidth]{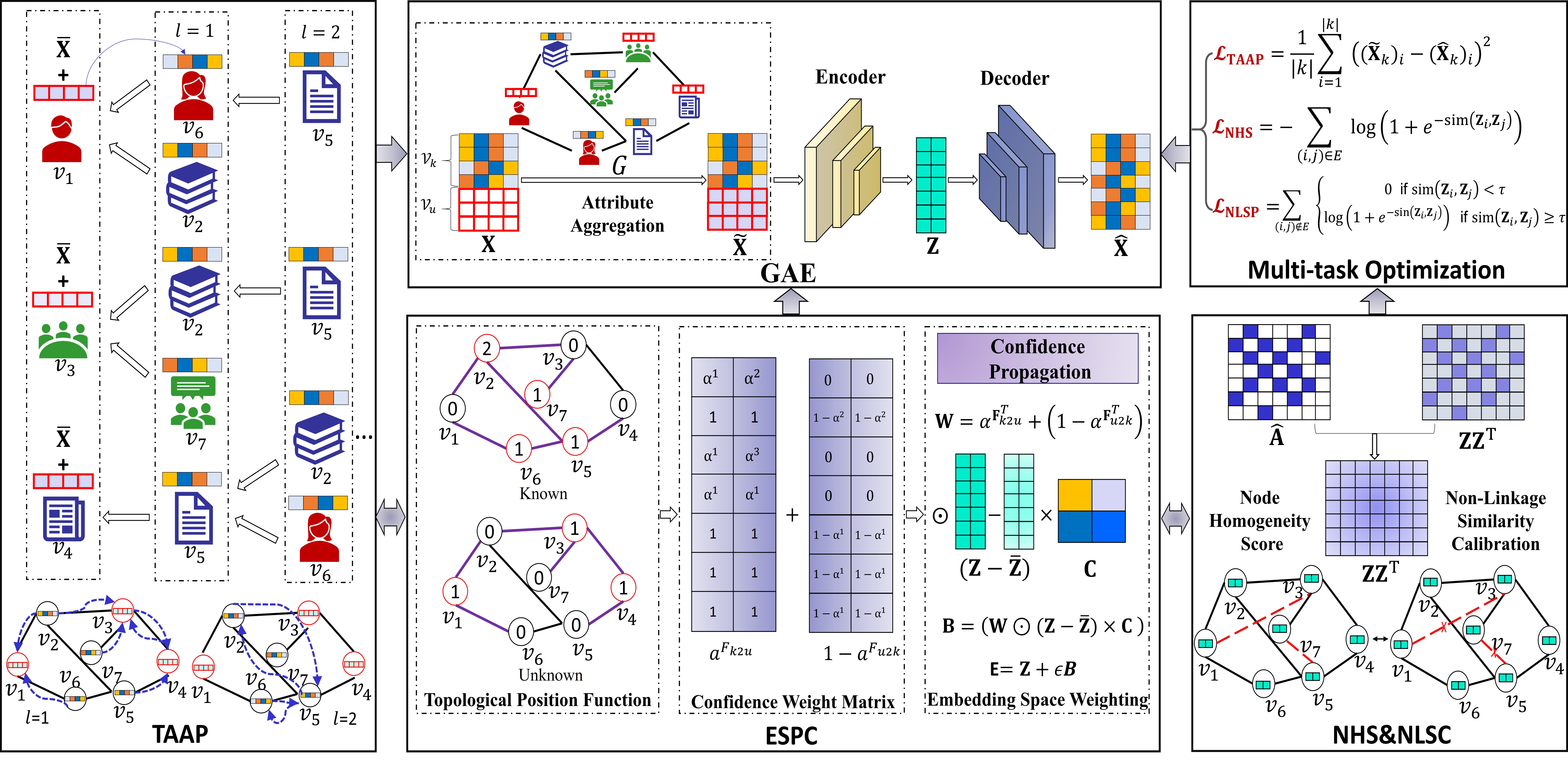}
\caption{The overall framework of TDAR. In the graph $G$ with missing node attributes, the attributes of nodes are first propagated through \textbf{TAAP} to aggregate known attribute features from different neighborhoods, obtaining refined features $\tilde{\mathbf X}$. Then, the  $\tilde{\mathbf X}$ and the normalized adjacency matrix $\hat{\mathbf A}$ are jointly fed into the GAE component to obtain  latent embeddings $\mathbf Z$ and reconstructed features $\hat{\mathbf X}$. Meanwhile, \textbf{ESPC} dynamically adjusts embeddings based on the confidence of relationships. In addition, \textbf{NHS} and \textbf{NLSC} scores are computed to provide auxiliary information or penalties to enhance the learning process.}
\label{fig1}
\end{figure*}

\subsection{Graph Auto-Encoder in AMGs}

In terms of recovering missing attributes, various methods such as \cite{chen2022learning, yoo2022accurate, xia2024attribute} have made corresponding improvements based on GAE, and TDAR is no exception. We next introduce the basic encoder, decoder, and optimization methods in TDAR.
\subsubsection{Encoder}
The encoder's primary function is to transform the input node features into a latent space represented by a continuous distribution. The encoding process for node feature attributes $\mathbf X$ and the symmetrically normalized adjacency matrix $\hat{\mathbf A}$ is described as:
\begin{equation}\label{e4}
    \mathbf Z = \text{GNN}(\mathbf X, \hat{\mathbf A}, \phi) = \sigma(\hat{\mathbf A} \cdot \text{ReLU}(\hat{\mathbf A} \cdot \mathbf X \phi_0) \cdot \mathbf{\phi}_1),
\end{equation}
where $\mathbf Z$ represents the latent space embedding, $\text{GNN}$ denotes a Graph Neural Network equipped with parameters $\mathbf {\phi}$ for encoding, and $\sigma(\cdot)$ is the Sigmoid activation function.

\subsubsection{Decoder}
The decoder, in contrast, aims to reconstruct the original or desired node attributes from the latent space. This process is formulated as:
\begin{equation}
    \mathbf {\hat X} = \text{MLP}(\mathbf Z, \theta) = \sigma(\text{ReLU}(\mathbf Z \cdot \theta_0) \cdot \theta_1),
\end{equation}
where $\mathbf {\hat X}$ denotes the reconstructed node features using a Multi-Layer Perceptron ($\text{MLP}$) driven by decoding parameters $\theta$.

\subsubsection{Objective Function}
In AMGs, GAE typically uses a customized objective function to optimize model parameters, focusing on the accurate reconstruction of node features and the integration of graph structure information. The objective function is defined as:
\begin{equation}\label{e31}
\mathcal{L}_{\text{Total}} = \mathcal{L}_\text{Reconstruction} + \mathcal{L}_\text{Regularization},
\end{equation}
where the reconstruction loss focuses on accurately rebuilding node attributes, and the regularization terms incorporate topological insights for handling missing attributes. Equation \eqref{e31} enhances TDAR's adaptability for multi-task learning, allowing the integration of various task-specific regularization components into a unified optimization framework. 

\section{Methodology}

The TDAR framework provides a comprehensive end-to-end approach for learning from graphs with missing attributes. It employs a self-supervised mechanism that integrates the processes of reconstructing missing node attributes and leveraging graph topology. Built on the GAE architecture, TDAR introduces four key strategies: TAAP enriches the initialization of missing attributes using topological insights; ESPC dynamically balances the embedding space; and NHS, alongside NLSC, collaboratively refines the identification of node relationships within the graph. Figure \ref{fig1} illustrates the framework of TDAR.

\subsection{Topology-Aware Attribute Propagation}

In the context of AMGs, nodes are influenced by their topological structure and connections with neighboring nodes. Leveraging these inherent structures and connection patterns aids in reconstructing missing attributes \cite{brin1998anatomy, rossi2022unreasonable, um2023confidence}. To address this, we introduce the TAAP, which utilizes graph topology to pre-fill missing attributes by identifying key dependencies. By providing a structured initialization based on the graph’s inherent connectivity, TAAP reduces noise during the learning process, allowing for more accurate attribute reconstruction and improved model stability.

Equation \eqref{e6} represents the general form of FP \cite{rossi2022unreasonable}, notably requiring no learnable parameters, thus making it suitable for pre-filling missing features in TAAP, enhancing performance in subsequent network framework learning. However, this method does not fully account for certain special cases; for instance, iterative convergence often requires a large number of iterations (typically 40), which can lead to excessive feature smoothing. Additionally, during the iteration process, the refined features of the known attribute nodes are not updated. 
To address these issues, our TAAP modifies the FP process by incorporating an additional matrix $\mathbf \Theta$: a small global average is introduced during the update process for unknown attributes to reduce the number of iterations, and known attribute features are allowed to participate in the iteration process, ultimately influencing the output of refined features. Therefore, the iterative process of Equation \eqref{e5} can be re-expressed as:
\begin{equation} \label{e7}
\begin{aligned}
\mathbf{X}^{(l+1)}&=\mathbf{X}^{(l)}-\begin{bmatrix}\mathbf{L}_{kk} & \mathbf{L}_{ku} \\ \mathbf{L}_{u k} & \mathbf{L}_{u u} \end{bmatrix} \mathbf{X}^{(l)} +\mathbf \Theta
\\&= \begin{bmatrix} \hat{\mathbf{A}}_{k k} & \hat{\mathbf{A}}_{k u} \\ \hat{\mathbf{A}}_{u k} & \hat{\mathbf{A}}_{u u}
\end{bmatrix}\begin{bmatrix}
\mathbf{X}_k^{(l)} \\
\mathbf{X}_u^{(l)}
\end{bmatrix} + \begin{bmatrix}
\alpha \overline {\mathbf X}_{k}^{(l)} \\
(1-\beta)\mathbf X_k^{(0)}
\end{bmatrix}.\\
\end{aligned}
\end{equation}

where $\overline{\mathbf{X}}_{k}^{(l)}$ represents the mean of the known attribute features of the $l$th layer, and $\alpha$ controls global propagation and $\beta$ controls the known reset, which are 0.05 and 0.1, respectively, in this paper. Based on this, the update process of Equation \eqref{e6} is re-expressed as:
\begin{equation}\label{e8}
\begin{aligned}
 \tilde{\mathbf X}_u^{(l+1)} &=  \hat{\mathbf A}_{uk} \mathbf X_k^{(l)} + \hat{\mathbf A}_{uu} \mathbf X_u^{(l)} + \alpha \overline {\mathbf X}_{k}^{(l)} \\
 \mathbf X_k^{(l+1)} &= \hat{\mathbf A}_{ku} \mathbf X_u^{(l)} + \hat{\mathbf A}_{kk} \mathbf X_k^{(l)}\\
\tilde{\mathbf X}_k^{(l+1)}  &=  \mathbf X_k^{(0)} + \beta \mathbf X_k^{(l+1)}.
\end{aligned}
\end{equation}
TAAP enhances the comprehensiveness of feature propagation and the smoothness of the reset process by introducing global information and known resets. Unlike FP, TAAP does not increase complexity during iterations, allowing for rapid pre-filling of unknown attributes (experimental results in Figure \ref{fig4} show that a maximum of 10 iterations is required) and using them alongside known attributes as the initial input features for GAE. Through TAAP, our method accurately infers underlying graph patterns to reconstruct missing attributes, thereby aiding in more precise attribute recovery.

\subsection{Embedding Space Propagation Confidence}

After feature pre-filling with TAAP, our TDAR can further obtain latent embeddings $\mathbf{Z} = \text{GNN}(\tilde{\mathbf X}, \hat{\mathbf A}, \phi)$ through the encoder defined in Equation \eqref{e4}. To balance the relationships between nodes in the latent space and optimize the signal-to-noise ratio, we introduced the ESPC strategy to distinguish the impact of different types of nodes on $\mathbf{Z}$. By using distance encoding \cite{zhang2018link, li2020distance}, we enhance the confidence of node connections, thereby improving the fidelity of information in the embedding space.

\subsubsection{Topological Position Function}
In addressing the propagation confidence of each node's position within the embedding space, we introduce two distance functions, $f_{\text{k2u}}(v)$ and $f_{\text{u2k}}(v)$, for capturing bidirectional influences between nodes with unknown and known attributes.

The function $f_{\text{k2u}}(v)$ computes the shortest path from an unknown attribute node $v \in V_u$ to the nearest known attribute node in $V_k$, thereby determining the confidence level of the unknown node:
\begin{equation}
f_{\text{k2u}}(v) = 
\begin{cases} 
0 & \text{if } v \in V_k \\
\min_{w \in V_k} d(v, w) & \text{if } v \in V_u
\end{cases},
\end{equation}
where $d(v, w)$ represents the shortest path distance using Breadth First Search (BFS) traversal \cite{um2023confidence}. 

Conversely, $f_{\text{u2k}}(v)$ quantifies the positional confidence of a known node by counting its direct connections to unknown nodes:
\begin{equation}
f_{\text{u2k}}(v) = 
\begin{cases} 
0 & \text{if } v \in V_u \\
|\{ w \in \mathcal{N}(v) : w \in V_u \}| & \text{if } v \in V_k
\end{cases},
\end{equation}
where $\mathcal{N}(v)$ denotes the neighborhood of node $v$. 

Using distance for unknown nodes ensures information propagates via the shortest path, reducing attenuation and noise. Using neighbor count for known nodes reflects their importance and influence in the network; nodes with more neighbors typically have higher connectivity and centrality.

\subsubsection{Confidence Weight Matrix}
In the context of graphs with missing attributes, accurately capturing the complex relationships between nodes is crucial. Our strategy, focused on balancing local and global information, results in dynamic latent embeddings. We introduce a weighting strategy to achieve this balance, factoring in both known and unknown attributes from neighboring nodes.

To modulate the influence of node relationships, we use a distance attenuation factor $\alpha \in (0,1)$. This factor helps in differentiating the confidence levels for unknown and known nodes. The weighting for unknown nodes $V_u$ decreases with increasing distance from known nodes $V_k$. In contrast, the weighting for known nodes increases with the number of connected unknown nodes. This dual approach balances distant and local influences, integrating both global and local node characteristics. The confidence weight matrix $\mathbf W\in \mathbb{R}^{N\times D}$ is defined as:
\begin{equation}
\mathbf W = \alpha^{\mathbf F_{\text{k2u}}^{T}} + (1 - \alpha^{\mathbf F_{\text{u2k}}^{T}}),
\end{equation}
where $\mathbf F_{\text{k2u}} \in \mathbb{R}^{N\times D}$ and $\mathbf F_{\text{u2k}} \in \mathbb{R}^{N\times D}$ expand $f_{\text{k2u}}$ and $f_{\text{u2k}}$.

\subsubsection{Embedding Space Weighting}
To elucidate the associations among various feature dimensions and potential node relationships, we examine the latent feature $\mathbf Z$, which is of dimension $D$. We compute a correlation matrix $\mathbf C \in \mathbb{R}^{D\times D}$:
\begin{equation}
\mathbf{C}_{ij}= \begin{cases}\frac{\sum_{v=1}^D\left(\mathbf{Z}_{iv}-\overline{\mathbf{Z}}_i\right)\left(\mathbf{Z}_{jv}-\overline{\mathbf{Z}}_j\right)}{\sqrt{\sum_{v=1}^D\left(\mathbf{Z}_{iv}-\overline{\mathbf{Z}}_i\right)^2 \sum_{v=1}^D\left(\mathbf{Z}_{jv}-\overline{\mathbf{Z}}_j\right)^2}} & \text { if } i \neq j \\ 0 & \text { if } i=j\end{cases},
\end{equation}
where $\bar{\mathbf Z}_i$ and $\bar{\mathbf Z}_j$ are the means of $\mathbf Z_i$ and $\mathbf Z_j$.

The final step combines the confidence weight matrix $\mathbf W$ with the correlation matrix $\mathbf C$ to produce $\mathbf B \in \mathbb{R}^{N\times D}$:
\begin{equation}
\mathbf B = \mathbf W \odot (\mathbf Z - \bar{\mathbf Z}) \times \mathbf C,
\end{equation}
where $\odot$ denotes element-wise multiplication.

The dynamic embeddings are then merged with the original embeddings to form a comprehensive representation:
\begin{equation}
\mathbf E = \mathbf Z + \epsilon \mathbf{B},
\end{equation}
where $\epsilon$ is a small constant. ESPC reduces the risk of information propagation errors by dynamically adjusting confidence levels in node relationships for different missing types, ensuring a precise and representative node influence in the embedding space. It is important to emphasize that $\mathbf{E}$ is used solely for calculating loss and is not updated in the original embeddings $\mathbf{E}$ within the GAE architecture.

\subsection{NHS and NLSC}

In the embedding space, assuming that similar attributes automatically imply proximity can introduce inaccuracies, particularly in the presence of incomplete or corrupted data \cite{zhu2020beyond, pei2020geom, lim2021large}. Ensuring the precise representation of node attributes and their graph connections is essential for effective learning. Based on the latent variables $\mathbf{E}$ obtained from ESPC, we introduce two key strategies: NHS evaluates the consistency among adjacent nodes, while NLSC corrects the misleading similarities among unconnected nodes. By leveraging the graph's topological relationships as prior knowledge, NHS and NLSC enhance the accuracy and stability of node embeddings, ensuring that structural integrity is maintained even in the presence of missing or corrupted data.

\subsubsection{Node Homogeneity Score}
Node homogeneity is evaluated based on the proximity of their embeddings. Nodes with close embeddings often share similar attributes or neighboring nodes, indicating homogeneity. We use cosine similarity to measure structural similarity between nodes. The NHS, considering both embedding and structure spaces, is defined as:
\begin{equation} 
\begin{aligned}
\text{NHS} = \sum_{(i,j) \in E} P(v_i \leftrightarrow v_j|\mathbf E) 
= \sum_{(i,j) \in E} \sigma(\text{sim}(\mathbf E_{i}, \mathbf E_{j})),
\end{aligned}
\end{equation}
where $\text{sim}(\mathbf E_{i}, \mathbf E_{j}) = \frac{\mathbf E_{i} \cdot \mathbf E_{j}}{\| \mathbf E_{i} \|_2 \times \| \mathbf E_{j} \|_2}$, $(i,j) \in E$ indicates an edge between nodes $i$ and $j$.

\subsubsection{Non-Linkage Similarity Calibration (NLSC)}

During embedding learning, the spatial distribution of original attributes may change, leading to non-linked nodes in the graph appearing similar in the embedding space without actual relationships. NLSC corrects these misleading similarities to ensure accurate representations in embeddings.

The similarity between non-linked nodes is calculated as $\sum_{(i,j) \notin E} \text{sim}(\mathbf E_i, \mathbf E_j)$. NLSC activates a calibration for node pairs without a connection in the graph if their embedding similarity exceeds a threshold $\tau$ (set to 0.2 in this paper):
% \begin{equation} 
% \text{NLSC} = \sum_{(i,j) \notin E} \begin{cases}0 & \text{if } \text{sim}(\mathbf Z_i, \mathbf Z_j) < \tau  \\  \sigma(\text{sim}(\mathbf Z_i, \mathbf Z_j)) & \text{if } \text{sim}(\mathbf Z_i, \mathbf Z_j) \geq \tau\end{cases},
% \end{equation}
\begin{equation} 
\begin{aligned}
\text{NLSC} &= P(v_i \nleftrightarrow v_j|\mathbf E) \\
&=\sum_{(i,j) \notin E} \begin{cases}0 & \text {if } \text{sim}(\mathbf E_i, \mathbf E_j) < \tau  \\  \sigma(\text{sim}(\mathbf E_i, \mathbf E_j)) & \text {if }\text{sim}(\mathbf E_i, \mathbf E_j) \geq \tau\end{cases}.
\end{aligned}
\end{equation}

While NHS ensures that the attribute similarities are faithfully represented among connected nodes, NLSC safeguards against false similarities among unconnected nodes. These mechanisms work in tandem to ensure that the embedding space accurately reflects both the attribute and structural properties of the graph, leading to more robust and reliable graph analysis.

\subsection{Multi-task Optimization Learning}

\subsubsection{Reconstruction Loss}
The refined attributes $\tilde{\mathbf X}$ generated by TAAP along with the symmetrically normalized  adjacency matrix $\hat{\mathbf A}$, are passed as input to GAE for the computation of the latent embeddings $\mathbf E$. Simultaneously, to ensure authenticity of the reconstruction, the reconstruction loss is calculated as the Mean Squared Error (MSE) between the known true attributes $\mathbf X_k$ and the known refined attributes $\tilde{\mathbf X}_k$:
\begin{equation}
\mathcal{L}_\text{TAAP} = \frac{1}{|k|} \sum_{i=1}^{|k|} \left((\tilde{\mathbf X}_{k})_{i} - (\hat{\mathbf X}_{k})_{i} \right)^2.
\end{equation}

\subsubsection{NHS Loss}
NHS effectively measures neighborhood consistency, making embedding learning dynamic and adaptive to neighborhood homogeneity. High NHS values suggest significant neighborhood homogeneity, while low values indicate the need for more attention in neighborhood information aggregation for embedding updates. The NHS-based loss function is given by:
\begin{equation}
\mathcal{L}_\text{NHS} = - \sum_{(i,j) \in E} \log(1 + e^{-\text{sim}(\mathbf E_{i}, \mathbf E_{j})}).
\end{equation}
\subsubsection{NLSC Loss}
NLSC effectively distinguishes among nodes that are similar in attributes but not actually related, enhancing the model's robustness against noisy and incomplete data. This calibration leads to improved generalization in downstream tasks. The loss function based on NLSC is defined as:
\begin{equation}
\mathcal{L}_\text{NLSC} = \sum_{(i,j) \notin E} \begin{cases}0 & \text{if } \text{sim}(\mathbf E_i, \mathbf E_j) < \tau  \\  \log(1 + e^{-\text{sim}(\mathbf E_{i}, \mathbf E_{j})}) & \text{if } \text{sim}(\mathbf E_i, \mathbf E_j) \geq \tau\end{cases}.
\end{equation}

Our TDAR framework integrates the loss functions of TAAP, NHS, and NLSC into Equation \eqref{e31}. The total loss function aims to optimize these strategies simultaneously, thereby improving overall performance, and is expressed as:
\begin{equation}
\mathcal{L}_{\text{Total}} = \underbrace{\mathcal{L}_{\text{TAAP}}}_{\text{Reconstruction}} + \underbrace{\lambda_1 \mathcal{L}_{\text{NHS}} + \lambda_2 \mathcal{L}_{\text{NLSC}}}_{\text{Regularization}},
\end{equation}
where $\lambda_1$ and $\lambda_2$ are hyperparameters that help achieve an effective balance between the different components.

\subsection{Complexity Analysis}

Analyzing the computational complexity of TDAR reveals its scalability and operational efficiency by evaluating the roles of nodes ($N$), edges ($M$), attribute features ($F$), and latent embedding dimensions ($D$).

The TAAP shows a complexity of $O(MF)$, indicating direct proportionality to the number of edges and features. The GAE adds complexity of $O((F+D)(M+ND))$, incorporating both feature and dimensionality effects alongside node and edge interactions. Computing the functions $f_{\text{k2u}}(v)$ and $f_{\text{u2k}}(v)$, which involve traversing nodes and their connections, yields $O(N+M)$. For the ESPC, processing the covariance matrix $\mathbf{C}$ and calculating node attributes result in $O(D^{2})$. The complexity of forming matrices $\mathbf{W}$ and $\mathbf{B}$ is $O(ND)$ each. For the NHS, the average computational complexity is $O(N \log N)$, necessitating analysis of each node and its neighbors. The NLSC potentially escalates to $O(N^2)$, due to exhaustive non-link evaluations across nodes.

Summarizing, TDAR's overall computational load integrates these components into a comprehensive complexity of $O\left((M+N)FD+N^2+ND^2+N\log N\right)$. This effectively simplifies to $O(N^2)$, situating TDAR as a competitive option relative to other sophisticated methodologies, as evidenced by existing research \cite{yoo2022accurate, peng2023multi}.

\section{Experimentation and Analysis}\label{Sec:IV}

This section addresses the following research questions:

\textbf{RQ1:} How does TDAR perform on downstream tasks such as feature reconstruction, classification, and clustering?

\textbf{RQ2:} What are the roles of each module in TDAR, as revealed by ablation studies?

\textbf{RQ3:} How does TDAR demonstrate robustness under varying input missing rates?

\textbf{RQ4:} What is the homogeneity capability of TDAR?

\textbf{RQ5:} How do changes in hyperparameters affect TDAR's performance?

\textbf{RQ6:} How visually effective is the quality of the learned graph representation learned by TDAR?

\textbf{RQ7:} How does TDAR's performance manifest in a specific case?

\subsection{Datasets}
In line with the benchmarks established in the domain, as highlighted in prior research by \cite{tu2022initializing, peng2023multi}, our analysis incorporates six highly recognized graph datasets: Cora, Citeseer, PubMed \cite{yang2016revisiting}, Amac, and Amap \cite{shchur2018pitfalls}, along with CS \cite{shchur2018pitfalls}. These datasets, enriched with essential graph components such as nodes, edges, and attribute features, provide a solid base for comprehensive analysis. Their varied graph structures and node attributes are widely employed in numerous graph-based tasks like representation learning, node classification, and clustering, highlighting their significance and utility for the objectives of our study. To assess the efficacy of our proposed TDAR method in contexts characterized by incomplete attribute data, we employ attribute masking—a technique inspired by \cite{chen2022learning, yoo2022accurate, peng2023multi}—to emulate the challenges of graphs with missing attribute information. Notably, the attribute features of Cora, CiteSeer, Amac, and Amap are binary, whereas the remaining datasets feature continuous attributes. A meticulous overview of the datasets is presented in Table \ref{tab1}, ensuring a clear and reproducible research method. 
\vspace{-2mm}
\begin{table}[htbp]
  \centering
  \caption{Data statistics.}
    \begin{tabular}{ccccccc}
    \toprule
    Dataset &{Type} & {Nodes} & {Edges} & {Features} & {Classes}  \\
    \hline
    Cora & Binary& 2,708 & 5,278 & 1,433 & 7  \\
    Citeseer &  Binary&3,327 & 4,228 & 3,703 & 6\\
    Amac & Binary&13,752 & 245,861 & 767 & 10 \\
    Amap &  Binary&7,650 & 119,081 & 745 & 8 \\
    PubMed &  Continuous&19,717 & 44,324 & 500 & 3 \\
    CS & Continuous& 18,333 & 81,894 & 6,805 & 15 \\
    \bottomrule
    \end{tabular}%
  \label{tab1}%
\end{table}%

\begin{table*}[htbp]\scriptsize
  \centering
  \caption{Comparative evaluation of TDAR and baseline methods on binary features for attribute reconstruction . Best results are indicated in ${\textbf{bold}_{\text{Improve}\%}}$, second best are {\underline{underlined}}.}
   \renewcommand\tabcolsep{3pt}
    \begin{tabular}{cccccccccccccc}
    \toprule
    \multirow{1}[2]{*}{Metric} & \multirow{1}[2]{*}{Method} & \multicolumn{3}{c}{Cora} & \multicolumn{3}{c}{Citeseer} & \multicolumn{3}{c}{Amac} & \multicolumn{3}{c}{Amap} \\
\cline{3-14}          &       & @10    & @20    & @50    & @10    & @20    & @50    & @10    & @20    & @50    & @10    & @20    & @50 \\
    \hline
    \multirow{9}[4]{*}{Recall} & NeighAggre & 0.0906  & 0.1413  & 0.1961  & 0.0511  & 0.0908  & 0.1501  & 0.0321  & 0.0593  & 0.1306  & 0.0329  & 0.0616  & 0.1361  \\
          & VAE   & 0.0887  & 0.1228  & 0.2116  & 0.0382  & 0.0668  & 0.1296  & 0.0255  & 0.0502  & 0.1196  & 0.0276  & 0.0538  & 0.1279  \\
          & GNN*   & 0.1350  & 0.1812  & 0.2972  & 0.0620  & 0.1097  & 0.2058  & 0.0273  & 0.0533  & 0.1278  & 0.0295  & 0.0573  & 0.1324  \\
          & GraphRNA & 0.1395  & 0.2043  & 0.3142  & 0.0777  & 0.1272  & 0.2271  & 0.0386  & 0.0690  & 0.1465  & 0.0390  & 0.0703  & 0.1508  \\
          & ARWMF  & 0.1291  & 0.1813  & 0.2960  & 0.0552  & 0.1015  & 0.1952  & 0.0280  & 0.0544  & 0.1289  & 0.0294  & 0.0568  & 0.1327  \\
          & SAT   & 0.1653  & 0.2345  & 0.3612  & 0.0811  & 0.1349  & 0.2431  & 0.0421  & 0.0746  & 0.1577  & 0.0427  & 0.0765  & 0.1635  \\
          & SVGA  & 0.1718  & 0.2486 & {\underline{0.3814}}  & 0.0943  & 0.1539  & {\underline{0.2782}}  & 0.0437  & 0.0769  & 0.1602  & {0.0446}  & {\underline{0.0798}}  & {\underline{0.1670}}  \\
         & ITR   & 0.1656  & 0.2372  & 0.3652  & 0.0972  & 0.1552  & 0.2679  & 0.0446  & 0.0780  & 0.1530  & 0.0434  & 0.0778  & 0.1635  \\
          & MATE  & {0.1731}  & 0.2460  & 0.3768  & 0.0994  & 0.1589  & 0.2716  & 0.0447  & 0.0782  & 0.1618  & 0.0442  & 0.0795  & 0.1649  \\
          & AIAE  & {\underline{0.1801}}  & \underline{0.2509}  & 0.3659  & {\underline{0.1047}}  & {\underline{0.1657}}  & 0.2717    & \underline{0.0461}  & \underline{0.0801} & \underline{0.1637} & {\underline{0.0449}}  & 0.0794  & 0.1652 \\
          & Ours  & ${\textbf{0.1855}_{\text{3.0}}}$ & ${\textbf{0.2598}_{\text{3.5}}}$ & ${\textbf{0.3913}_{\text{2.6}}}$ & ${\textbf{0.1062}_{\text{1.4}}}$ & ${\textbf{0.1683}_{\text{1.6}}}$ & ${\textbf{0.2852}_{\text{2.5}}}$ & ${\textbf{0.0474}_{\text{2.8}}}$ & ${\textbf{0.0825}_{\text{3.0}}}$ & ${\textbf{0.1680}_{\text{2.6}}}$ & ${\textbf{0.0469}_{\text{4.5}}}$ & ${\textbf{0.0827}_{\text{3.6}}}$ & ${\textbf{0.1710}_{\text{2.4}}}$ \\
    \hline
    \multirow{9}[2]{*}{nDCG} & NeighAggre & 0.1217  & 0.1548  & 0.1850  & 0.0823  & 0.1155  & 0.1560  & 0.0788  & 0.1156  & 0.1923  & 0.0813  & 0.1196  & 0.1998  \\
          & VAE   & 0.1224  & 0.1452  & 0.1924  & 0.0601  & 0.0839  & 0.1251  & 0.0632  & 0.0970  & 0.1721  & 0.0675  & 0.1031  & 0.1830  \\
          & GNN*   & 0.1736  & 0.2076  & 0.2702  & 0.1026  & 0.1423  & 0.2049  & 0.0671  & 0.1027  & 0.1824  & 0.0705  & 0.1082  & 0.1893  \\
          & GraphRNA & 0.1934  & 0.2362  & 0.2938  & 0.1291  & 0.1703  & 0.2358  & 0.0931  & 0.1333  & 0.2155  & 0.0959  & 0.1377  & 0.2232  \\
          & ARWMF  & 0.1824  & 0.2182  & 0.2776  & 0.0859  & 0.1245  & 0.1858  & 0.0694  & 0.1053  & 0.1851  & 0.0727  & 0.1098  & 0.1915  \\
          & SAT   & 0.2250  & 0.2723  & 0.3394  & 0.1385  & 0.1834  & 0.2545  & 0.1030  & 0.1463  & 0.2346  & 0.1047  & 0.1498  & 0.2421  \\
          & SVGA  & 0.2381  & 0.2894  & {\underline{0.3601}}  & 0.1579  & 0.2076  & 0.2892  & 0.1068  & 0.1509  & 0.2397  & 0.1084  & 0.1549  & {\underline{0.2472}}  \\
          & ITR   & 0.2288  & 0.2770  & 0.3448  & 0.1645  & 0.2129  & 0.2870  & 0.1086  & 0.1562  & 0.2415  & 0.1069  & 0.1526  & 0.2440  \\
          & MATE  & 0.2373  & 0.2861  & 0.3550  & 0.1708  & 0.2204  & 0.2950  & 0.1090  & 0.1535  & 0.2424  & 0.1086  & {0.1553}  & 0.2465  \\
            & AIAE  & {\underline{0.2487}}  & \underline{0.2958}  & 0.3570  & {\underline{0.1774}}  & {\underline{0.2285}}  & \underline{0.2983}    & \underline{0.1126}  & \underline{0.1576} & \underline{0.2464} & {\underline{0.1096}}  & {\underline{0.1556}}  & 0.2468 \\
          & Ours  & ${\textbf{0.2582}_{\text{3.8}}}$ & ${\textbf{0.3079}_{\text{4.1}}}$ & ${\textbf{0.3783}_{\text{5.1}}}$ & ${\textbf{0.1799}_{\text{1.4}}}$ & ${\textbf{0.2317}_{\text{1.4}}}$ & ${\textbf{0.3086}_{\text{3.4}}}$ & ${\textbf{0.1150}_{\text{2.1}}}$ & ${\textbf{0.1614}_{\text{2.4}}}$ & ${\textbf{0.2521}_{\text{2.3}}}$ & ${\textbf{0.1133}_{\text{3.4}}}$ & ${\textbf{0.1606}_{\text{3.2}}}$ & ${\textbf{0.2541}_{\text{2.8}}}$ \\
    \bottomrule
    \end{tabular}%
  \label{tab2}%
\end{table*}%

\begin{table}[htbp]\scriptsize
  \centering
  \caption{Evaluation on continuous features for attribute reconstruction.}
   \renewcommand\tabcolsep{3pt}
    \begin{tabular}{ccccc}
    \toprule
    \multirow{2}[4]{*}{Method} & \multicolumn{2}{c}{PubMed} & \multicolumn{2}{c}{CS} \\
\cmidrule{2-5}          & RMSE ($\downarrow$)  & CORR ($\uparrow$)   & RMSE ($\downarrow$) & CORR ($\uparrow$)  \\
    \midrule
    NeighAgg & 0.0186  & -0.2133  & 0.0952  & -0.2279  \\
    VAE   & 0.0170  & -0.0236  & 0.0863  & -0.0237  \\
    GGN  & 0.0168  & -0.0010  & 0.0850  & 0.0179  \\
    GraphSAGE  & 0.0171  & 0.0054  & 0.0854  & 0.0154  \\
    GAT  & 0.0174  & 0.0130  & 0.0852  & 0.0187  \\
    GraphRNA & 0.0172  & -0.0352  & 0.0897  & -0.1052  \\
    ARWMF & 0.0165  & \underline{0.0434}  & 0.0827  & 0.0710  \\
    SAT   & 0.0165  & {0.0378}  & \underline{0.0820}  & \underline{0.0958}  \\
    SVGA  & 0.0166  & 0.0280  & 0.0824  & 0.0740  \\
    ITR   & \underline{0.0164}  & 0.0324  & 0.0832  & 0.0543  \\
    MATE & {0.0165}  & 0.0425 &{0.0830}  &0.0735  \\
    AIAE&{0.0166}  & 0.0425 &{0.0827}  &0.0760  \\
    Ours  & ${\textbf{0.0160}_{\text{2.4}}}$  & ${\textbf{0.0943}_{\text{117}}}$  & ${\textbf{0.0812}_{\text{1.0}}}$  &${\textbf{0.1033}_{\text{7.8}}}$  \\
    \bottomrule
    \end{tabular}%
  \label{tab_21}%
\end{table}%

\subsection{Baselines}
In our experiments, we benchmarked TDAR against various established baseline methods, as detailed in the following:

NeighAgg \cite{csimcsek2008navigating}: Utilizes mean pooling to aggregate adjacent node features, focusing on one-hop neighbors.
VAE \cite{kingma2013auto}: A generative model capturing latent instance representations. In our context, it aids in reconstructing absent node attributes.
GNN*: Represents the top models from renowned graph neural networks such as GCN \cite{kipf2016semi}, GraphSAGE \cite{hamilton2017inductive}, and GAT \cite{velivckovic2017graph}, prevalent across diverse sectors.
GraphRNA \cite{huang2019graph} \& ARWMF \cite{chen2019attributed}: Both are contemporary representation learning methodologies oriented towards feature generation.
SAT \cite{chen2022learning}: Advocates for a shared latent realm to nurture independent autoencoders, bridging features and graph structures.
SVGA \cite{yoo2022accurate}: A leading-edge model which brings Gaussian Markov random fields into play, delineating the feature space across attributes and graph composition.
ITR \cite{tu2022initializing}: Offers an integrated solution that taps into the graph's inherent topology, facilitating adaptive mending of missing attributes.
MATE \cite{peng2023multi}: Rolls out a dual-perspective framework, championing contrastive learning by leveraging attribute and topology view augmentations.
AIAE \cite{xia2024attribute}: Employs a dual encoder based on knowledge distillation and a multi-scale decoder with masking to improve the accuracy of imputation.
AMGC \cite{tu2024attribute}: Improves clustering performance on AMGs by alternately promoting clustering and imputation within a unified framework.

\subsection{Experimentation Settings}
\textbf{Dataset Split} In this study, following the settings of \cite{yoo2022accurate}, we fixed the random seed number at 72, while partitioned and masked the dataset into 40\% for training, 10\% for validation, and 50\% for testing. Of these, 40\% of the training set consisted of known nodes, with the remainder being unknown nodes with attribute masking.

\textbf{Hyperparameters} For the optimization technique, Adam was chosen \cite{kingma2014adam}. Learning rates of 0.001 for the Cora and Citeseer datasets were set, employing a encoder and decoder structure, both of which were single-layer MLPs, with a 0.8 dropout rate. Latent layer count stood at 256, persisting for 1000 runs. Rates for Amac, Amcp, PubMed and CS followed at 0.01, with a binary-tier GCN-MLP for encoder-decoder framework, amending dropout to 0.2, and an endurance of 400 epochs. Hyperparameters maintained across the panel were: $\lambda_1=0.1$, $\lambda_2=0.1$, $l=10$, $\epsilon=0.01$, and $\alpha=0.9$, provided no special-case study.

\textbf{Downstream Metrics} To assess the quality of \textbf{attribute reconstruction}, we introduced Recall@K and nDCG@K metrics at fixations k of 10, 20, and 50. For continuous features, we used RMSE and CORR. \cite{chen2022learning}. For the \textbf{node classification}, we trained a basic MLP as the classifier and conducted a five-fold cross-validation on the reconstructed samples, respectively \cite{chen2022learning}. The performance was gauged by the classification Accuracy. In terms of \textbf{node clustering}, all node embeddings from the latent space were clustered using the K-means \cite{lloyd1982least}. Evaluation metrics included Accuracy (ACC), NMI, ARI, and F1 Score (F1) \cite{peng2023multi}.

\subsection{Performance on Downstream Tasks (\textbf{RQ1})}
To fully validate TDAR, we compared with contrasting methods on three downstream tasks, including attribute feature reconstruction, node classification, and node clustering.

\subsubsection{Attribute Feature Reconstruction}
In our comparative experiments into the reconstruction of features for AMGs, we employed a systematic process to evaluate and compare a variety of cutting-edge technologies. As demonstrated in Tables \ref{tab2} and \ref{tab_21}, we selected several widely recognized classic methods, as well as some of the current leading algorithms such as SAT, SVGA, ITR, MATE, and AIAE.

Preliminary analysis indicates that our proposed TDAR method consistently excels in benchmark tests across all datasets considered. In contrast, methods like NeighAggre and VAE underperform, with this disparity likely stemming from their inability to adequately mine and represent the key characteristics of data during the encoding stage. Although GNN*, GraphRNA, and ARWMF have shown some performance improvements in certain respects, they still fall short of expectations. Their limitations are mainly due to their graph neural network-based approach to learning data's topological features but showing clear deficiencies in modules specifically designed for graphs with missing attributes.

On the other hand, state-of-the-art methods such as SVGA, ITR, MATE and AIAE have exhibited outstanding results in multiple experiments. These methods, through in-depth exploration and careful design, focus on addressing the core issues and challenges associated with graphs with missing attributes, and effectively mine the intrinsic properties of data, thereby achieving impressive performance gains. However, it is worth noting that when comparing all these techniques collectively, our developed TDAR method distinctly stands out due to its comprehensive optimization strategy.  TDAR achieved the best results across all metrics on every dataset, with an average improvement of around 3\%. The TDAR leverages the TAAP technique for refined attribute pre-filling, uses nodes' topological credibility and similarity as weights to update embeddings, and applies precise feature reconstruction supervision, offering a comprehensive performance improvement in multi-dimensional learning.

\begin{table}[htbp]\scriptsize
  \centering
  \caption{Comparison of node classification accuracy between TDAR and baselines using reconstructed features. ``o.o.m." indicates Out-of-Memory.}
   \renewcommand\tabcolsep{3.5pt}
    \begin{tabular}{ccccccc}
    \toprule
     Method  & Cora  &  Citeseer  & Amac & Amap& PubMed& CS \\
    \hline
    NeighAggre & 0.6248  & 0.5593  & 0.8365  & 0.8846 &0.5150 & 0.7562 \\
    VAE   & 0.2826  & 0.2551  & 0.3747  & 0.2598 &0.3854& 0.4008   \\
    GNN*  & 0.4525  & 0.2688  & 0.4034  & 0.3789 &0.4013 & 0.2317  \\
    GraphRNA & 0.7581  & 0.6320  & 0.6968  & 0.8407& 0.6035 & 0.7710    \\
    ARWMF & 0.7769  & 0.2267  & 0.5608  & 0.4675 & 0.6180 & 0.2320 \\
    SAT   & 0.7644  & 0.6010  & 0.7410  & 0.8762 & 0.4618 & 0.7672  \\
    SVGA  & {\underline{0.8498}}  & 0.6805  & 0.8725  & 0.9077 & \underline{0.6227} & \underline{0.8293}  \\
    ITR   & 0.8143  & 0.6715  & 0.8388  & 0.9075  & o.o.m.& o.o.m. \\
    MATE  & 0.8327  & 0.6805  & {\underline{0.8827}}  & {\underline{0.9168}} & o.o.m.& o.o.m.  \\
    AIAE & 0.8401 & \underline{0.6833} & 0.8591 & 0.9094 & o.o.m. & o.o.m.  \\
    Ours  & ${\textbf{0.8597}_{\text{1.2}}}$ & ${\textbf{0.6890}_{\text{0.8}}}$ &${\textbf{0.9047}_{\text{2.5}}}$ & ${\textbf{0.9294}_{\text{1.4}}}$ & ${\textbf{0.8226}_{\text{32.1}}}$&${\textbf{0.9086}_{\text{9.5}}}$ \\
    \bottomrule
    \end{tabular}%
  \label{tab3}%
\end{table}%

\subsubsection{Node Classification}

When assessing the reconstructed features, it is crucial to look beyond the immediate reconstruction results and consider their performance on downstream tasks. This helps gauge the true utility of the reconstruction, revealing how well the reconstructed features can support subsequent analyses and practical applications.

Specifically, we used the reconstructed features with an MLP as a classifier for the node classification task, and the results are shown in Table \ref{tab3}. Overall, it is clear that our TDAR method consistently outperforms the other methods across all four datasets. This indicates that TDAR's capability in attribute reconstruction not only preserves essential features but also enhances the discriminative power necessary for node classification tasks. Upon examining the specifics, NeighAggre and VAE exhibit subpar performance across all datasets, suggesting a deficiency in reconstructing features critical for classification. In contrast, while GraphSage, GAT, and GCN offer better metrics than the former two, they still fall short of the state-of-the-art methods. This may highlight the challenges they face in balancing effective attribute reconstruction with subsequent classification accuracy. Methods such as GraphRNA, SAT, SVGA, ITR, and MATE display commendable accuracy, coming close to that of TDAR. Their success can be attributed to their meticulous approach to attribute reconstruction, which in turn, has somewhat enhanced the overall accuracy of the classification tasks.

\subsubsection{Node Clustering}

\begin{table}[htbp]\scriptsize
  \centering
  \caption{Comparison of node clustering accuracy between TDAR and baselines using reconstructed features.}
  %\caption{Comparison of node clustering accuracy between TDAR and baselines using latent features.}
   \renewcommand\tabcolsep{3.5pt}
    \begin{tabular}{cccccccccc}
    \toprule
    Dataset & Metric  & GNN*   & SAT   & SVGA  & ITR   & MATE &AMGC & Ours \\
    \hline
    \multirow{4}[1]{*}{Cora} & ACC   & 0.3922  & 0.4125 & 0.4217 & 0.3552 & 0.6019 &\underline{0.6665}& ${\textbf{0.7097}_{\text{6.5}}}$ \\
          & NMI   & 0.2483  & 0.2534 & 0.2900  & 0.1745 & 0.4144 & \underline{0.4799}&${\textbf{0.5293}_{\text{10.3}}}$ \\
          & ARI    & 0.1794  & 0.1817 & 0.1744 & 0.0318 & 0.3630 & \underline{0.4340}&${\textbf{0.4805}_{\text{4.7}}}$ \\
          & F1   & 0.3502  & 0.3607 & 0.3863 & 0.3112 & {0.5022}&\underline{0.6102}&${\textbf{0.6452}_{\text{5.7}}}$ \\
              \hline
    \multirow{4}[0]{*}{Citeseer} & ACC   & 0.3375  & 0.3372 & 0.4491 & 0.3339 & 0.5774&\underline{0.6092} & ${\textbf{0.6369}_{\text{4.5}}}$ \\
          & NMI   & 0.1684  & 0.1115 & 0.2407 & 0.1609 & 0.2963 & \underline{0.3293}&${\textbf{0.3526}_{\text{7.0}}}$ \\
          & ARI  & 0.0817  & 0.0875 & 0.1492 & 0.0366 & 0.2934 & \underline{0.3373}&${\textbf{0.3658}_{\text{8.4}}}$ \\
          & F1    & 0.2779 & 0.3184 & 0.4280 & 0.3403 & 0.5348 & \underline{0.5733}&${\textbf{0.5902}_{\text{2.9}}}$ \\
       \hline
    \multirow{4}[1]{*}{Amac} & ACC   & 0.4369  & 0.4930 & 0.3426 & 0.3490 & {\underline{0.5302}}&- & ${\textbf{0.6392}_{\text{20.6}}}$ \\
          & NMI   & 0.1523 & 0.4304 & 0.0554 & 0.2624 & {\underline{0.4786}} & -&${\textbf{0.5348}_{\text{11.7}}}$ \\
          & ARI   & 0.1868 & {\underline{0.3518}} & 0.0849 & 0.1366 & 0.3212 & -&${\textbf{0.4802}_{\text{36.9}}}$ \\
          & F1   & 0.1778 & 0.3367 & 0.1216 & 0.2539 & {\underline{0.4901}} & -&${\textbf{0.5487}_{\text{12.0}}}$ \\
    \hline
    \multirow{4}[2]{*}{Amap} & ACC  & 0.4248 & 0.5471 & 0.3511 & 0.3574 & {\underline{0.5846}} &-& ${\textbf{0.7885}_{\text{34.9}}}$ \\
          & NMI   & 0.1565 & {\underline{0.5146}} & 0.1305 & 0.2249 & 0.5077 & -&${\textbf{0.7070}_{\text{37.4}}}$ \\
          & ARI    & 0.1454 & {\underline{0.3825}} & 0.0820 & 0.0298 & 0.2764 & -&${\textbf{0.6104}_{\text{59.6}}}$ \\
          & F1    & 0.2507 & 0.4734 & 0.1852 & 0.3241 & {\underline{0.5127}} & -&${\textbf{0.7290}_{\text{42.2}}}$ \\
    \bottomrule
    \end{tabular}%
  \label{tab4}%
\end{table}%
Table \ref{tab4} meticulously compares our TDAR method with other baseline methods in node clustering tasks. The experiments utilized K-means clustering of latent embeddings $\mathbf Z$ to form clusters. The results conclusively show that TDAR consistently outperforms competing methods across all four datasets. This indicates that TDAR's proficiency in attribute reconstruction not only preserves fundamental features but also significantly enhances its capability to effectively organize nodes into coherent clusters.
A closer examination reveals that although graph learning methods like GCN, GAT, and GraphSAGE are highly acclaimed in the field, they often fall short in the node clustering paradigm across most datasets. Their relatively lower performance metrics indicate inherent challenges in transforming high-quality feature reconstructions into effective clustering outcomes. In contrast, techniques such as SAT, SVGA, ITR, and MATE exhibit more robust performances, benefiting from superior feature reconstruction strategies that better facilitate the clustering process.
Moreover, AMGC, which focuses on AMG clustering, represents the forefront of current methodologies. However, since AMGC has not been open-sourced, it has not been tested on the Amac and Amap datasets and comparisons have been restricted to the Cora and Citeseer datasets. Despite this limitation, our TDAR method still achieves an average improvement of 2-10\% over AMGC, underscoring its effectiveness and the advanced capabilities of our feature reconstruction process in producing more refined and cohesive clustering.

\begin{table*}[htbp]\scriptsize
  \centering
  \caption{Ablation experiments for attribute reconstruction.}
   \renewcommand\tabcolsep{3.5pt}
    \begin{tabular}{cccccccccccccc}
    \toprule
    \multirow{1}[2]{*}{Metric} & \multirow{1}[2]{*}{Method} & \multicolumn{3}{c}{Cora} & \multicolumn{3}{c}{Citeseer} & \multicolumn{3}{c}{Amac} & \multicolumn{3}{c}{Amap}\\
\cline{3-14}          &       & @10    & @20    & @50    & @10    & @20    & @50    & @10    & @20    & @50    & @10    & @20    & @50 \\
    \hline
    \multirow{6}[1]{*}{Recall} & (1) Baseline (GAE) & 0.1651  & 0.2315  & 0.3554  & 0.0929  & 0.1464  & 0.2556  & 0.0401  & 0.0715  & 0.1529  & 0.0412  & 0.0742  & 0.1590  \\
          & (2) w/ FP & 0.1687  & 0.2364  & 0.3600  & 0.0935  & 0.1489  & 0.2568  & 0.0414  & 0.0730  & 0.1549  & 0.0426  & 0.0757  & 0.1620  \\
          & (3) w/ TAAP & 0.1740  & 0.2439  & 0.3728  & 0.0951  & 0.1497  & 0.2586  & 0.0447  & 0.0776  & 0.1615  & 0.0446  & 0.0791  & 0.1672  \\
          & (4) w/ TAAP ESPC NHS & \underline{0.1824}  & \underline{0.2572}  & \underline{0.3884}  & 0.1012  & \underline{0.1621}  & \underline{0.2757}  & 0.0445  & 0.0775  & 0.1607  & 0.0447  & 0.0792  & 0.1657  \\
          & (5) w/ TAAP ESPC NLSC & 0.1717  & 0.2381  & 0.3524  & \underline{0.1018}  & 0.1608  & 0.2740  & \underline{0.0450}  & \underline{0.0787}  & \underline{0.1628}  & \underline{0.0450}  & \underline{0.0801}  & \underline{0.1679}  \\
          & (6) TDAR & \textbf{0.1855} & \textbf{0.2598} & \textbf{0.3913} & \textbf{0.1062} & \textbf{0.1683} & \textbf{0.2852} & \textbf{0.0474} & \textbf{0.0825} & \textbf{0.1680} & \textbf{0.0469} & \textbf{0.0827} & \textbf{0.1710} \\
    \hline
    \multirow{6}[1]{*}{nDCG} & (1) Baseline (GAE) & 0.2313  & 0.2761  & 0.3410  & 0.1587  & 0.2037  & 0.2754  & 0.0986  & 0.1406  & 0.2273  & 0.1005  & 0.1444  & 0.2345  \\
          & (2) w/ FP & 0.2364  & 0.2817  & 0.3472  & 0.1614  & 0.2078  & 0.2787  & 0.1012  & 0.1434  & 0.2305  & 0.1031  & 0.1472  & 0.2387  \\
          & (3) w/ TAAP & 0.2458  & 0.2927  & 0.3605  & 0.1639  & 0.2096  & 0.2812  & 0.1093  & 0.1532  & 0.2423  & 0.1084  & 0.1542  & 0.2473  \\
          & (4) w/ TAAP ESPC NHS & \underline{0.2531}  & \underline{0.3035}  & \underline{0.3732}  & \underline{0.1733}  & \underline{0.2242}  & \underline{0.2988}  & 0.1093  & 0.1533  & 0.2419  & 0.1087  & 0.1544  & 0.2462  \\
          & (5) w/ TAAP ESPC NLSC & 0.2390  & 0.2829  & 0.3439  & 0.1721  & 0.2214  & 0.2960  & \underline{0.1101}  & \underline{0.1547}  & \underline{0.2441}  & \underline{0.1091}  & \underline{0.1555}  & \underline{0.2485}  \\
          & (6) TDAR & \textbf{0.2582} & \textbf{0.3079} & \textbf{0.3783} & 0.1798  & \textbf{0.2317} & \textbf{0.3086} & \textbf{0.1150} & \textbf{0.1614} & \textbf{0.2521} & \textbf{0.1133} & \textbf{0.1606} & \textbf{0.2541} \\
    \bottomrule
    \end{tabular}%
  \label{tab5}%
\end{table*}%

\begin{table}[htbp]\scriptsize
  \centering
  \caption{Ablation experiments for node classification.}
   \renewcommand\tabcolsep{3.5pt}
    \begin{tabular}{ccccc}
    \toprule
     Method  & Cora  &  Citeseer  & Amac  & Amap \\
    \hline
    (1) Baseline (GAE) & 0.7483  & 0.6274  & 0.8383  & 0.8756  \\
    (2) w/ FP & 0.7612  & 0.6359  & 0.8625  & 0.8906  \\
    (3) w/ TAAP & 0.7791  & 0.6375  & 0.9003  & 0.9266  \\
    (4) w/ TAAP ESPC NHS & 0.8209  & 0.6690  & 0.9020  & 0.9277  \\
    (5) w/ TAAP ESPC NLSC & \underline{0.8388}  & \underline{0.6720}  & \underline{0.9024}  & \underline{0.9283}  \\
    (6) TDAR & \textbf{0.8597} & \textbf{0.6950} & \textbf{0.9047} & \textbf{0.9294} \\
    \bottomrule
    \end{tabular}%
  \label{tab51}%
\end{table}%

\begin{figure}[t]
\centering
\includegraphics[width=0.5\textwidth]{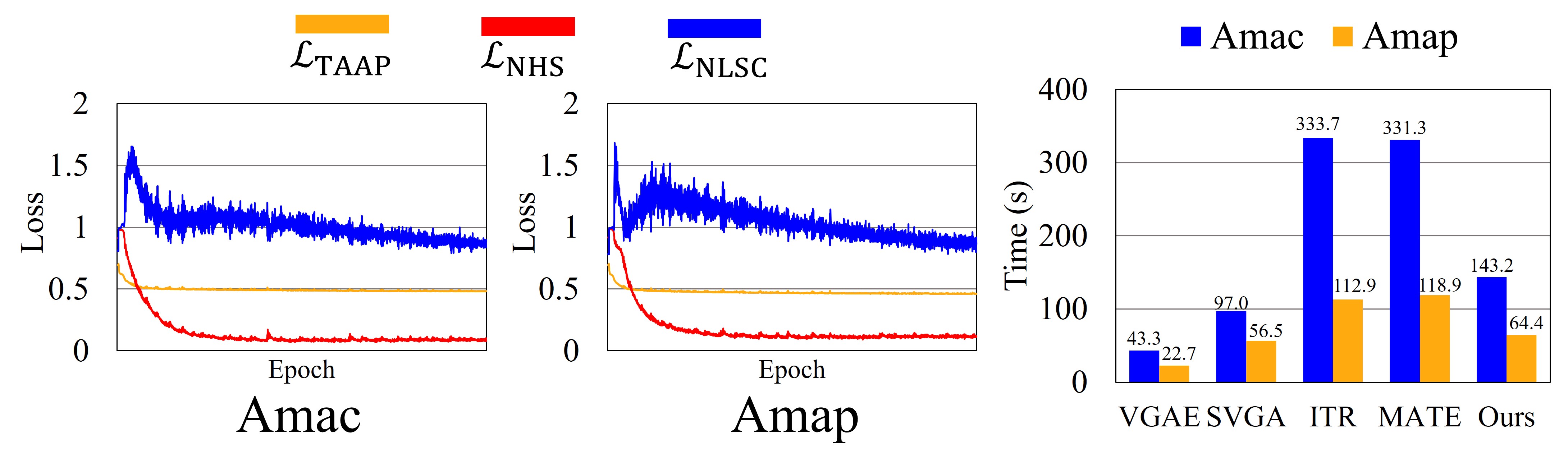}
\caption{Comparison of convergence status and training time.}
\label{figa1}
\end{figure}

\subsection{Ablation Studies (\textbf{RQ2})}

\textbf{Component Ablations} To evaluate the effectiveness of each component in TDAR, we conducted ablation experiments on two sub-tasks: attribute reconstruction and node classification, as shown in Tables \ref{tab5} and \ref{tab51}. The ablation study provided five variants in addition to TDAR. Among them, (1) is the baseline method, where we selected a basic GAE architecture and included only the reconstruction loss; (2) indicates the addition of FP to the baseline to demonstrate the effectiveness of pre-filling; (3) represents the addition of TAAP to the baseline, where TAAP introduces global mean and known attribute updates based on FP; (4) includes TAAP, ESPC, and NHS; (5) includes TAAP, ESPC, and NLSC; and (6) is our TDAR method. We found that TDAR consistently outperforms the other variants across all datasets.

Further analysis of individual ablation variants revealed the following:

\begin{itemize}
\item The GAE baseline performs the worst. The results improve with the addition of FP in variant (2), and replacing FP with TAAP in variant (3) further improves performance, indicating that feature pre-filling is effective in AMG learning, and TAAP enhances FP to further improve prediction results.
\item ESPC, as a prerequisite for NHS and NLSC, further enhances model performance when added. However, we observed that replacing NHS with NLSC in variant (5) slightly decreases the attribute reconstruction scores on the Cora and Citeseer datasets, indicating that their roles in the TDAR framework are complementary.
\item In variant (3), TAAP has a more significant impact on attribute reconstruction than on node classification, while in variants (4) and (5), the addition of ESPC, NHS, and NLSC significantly improves node classification performance. This indicates that different modules have unique advantages in different tasks.
\item Excluding TAAP, ESPC, NHS, and NLSC results in a significant performance drop for TDAR, highlighting that these components work synergistically to achieve optimal efficiency.
\end{itemize}

\textbf{Convergency} In this study, we conducted a detailed evaluation of the convergence of three loss functions, as shown in Figure \ref{figa1}. For clarity and to illustrate trends, we normalized all loss function values in our analysis. Our analysis indicates that these loss functions exhibit stable convergent behavior at different stages of training. Subsequently, we thoroughly assessed the time expenditure of TDAR. Compared to leading benchmark models such as MATE and ITR, TDAR demonstrates significant advantages in time efficiency. Moreover, compared to SVGA, TDAR shows a slight increase in time expenditure, suggesting that our model can offer competitive performance with a limited sacrifice in time efficiency.

\begin{figure*}[t]
\centering
\includegraphics[width=0.98\textwidth]{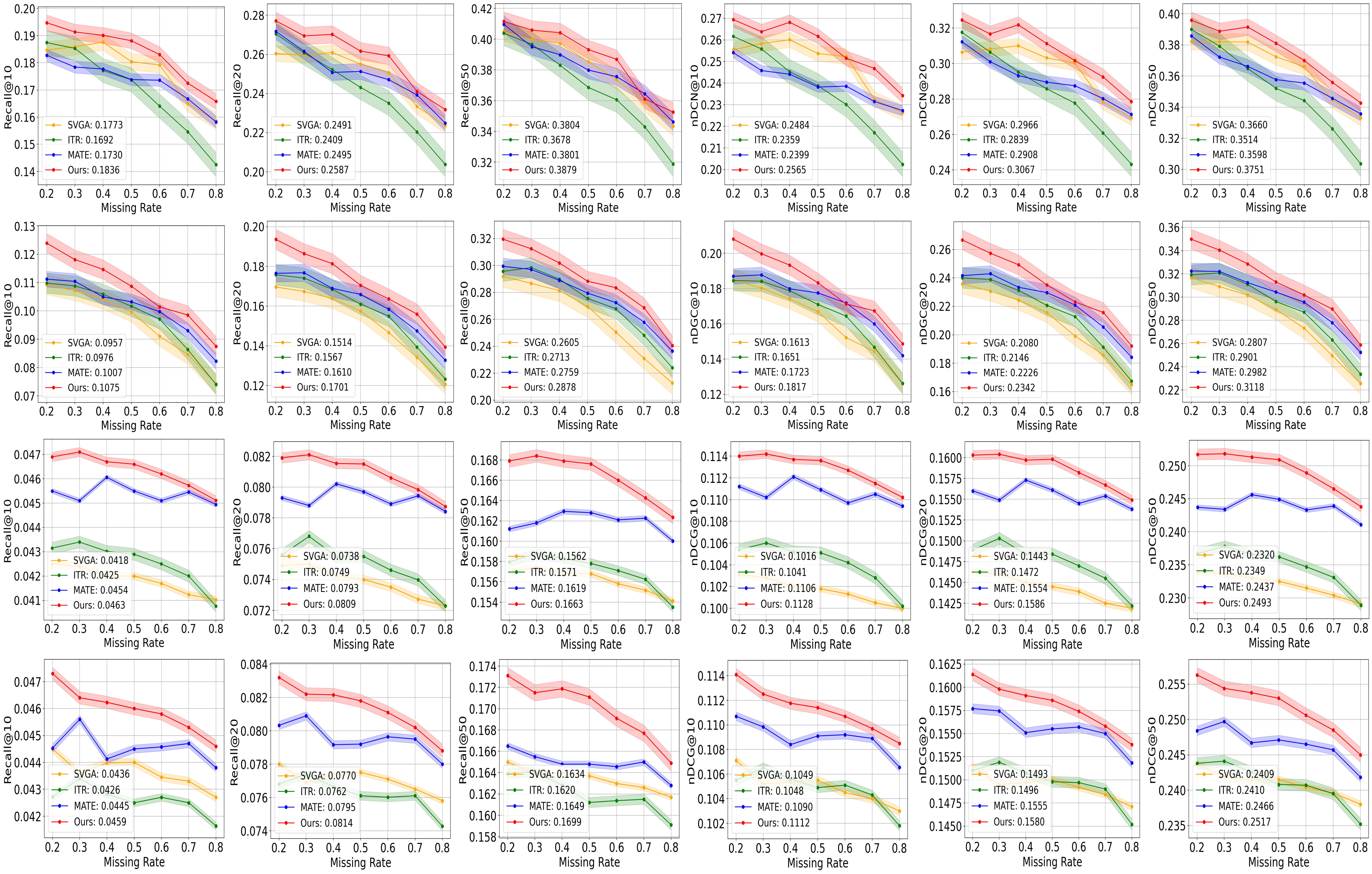}
\caption{Performance comparison under different missing rates. The legend counts the average of all missing rates. Datasets Cora, Citeseer, Amac and Amap are arranged by row from top to bottom.}
\label{fig2}
\end{figure*}

\begin{figure}[t]
\centering
\includegraphics[width=0.45\textwidth]{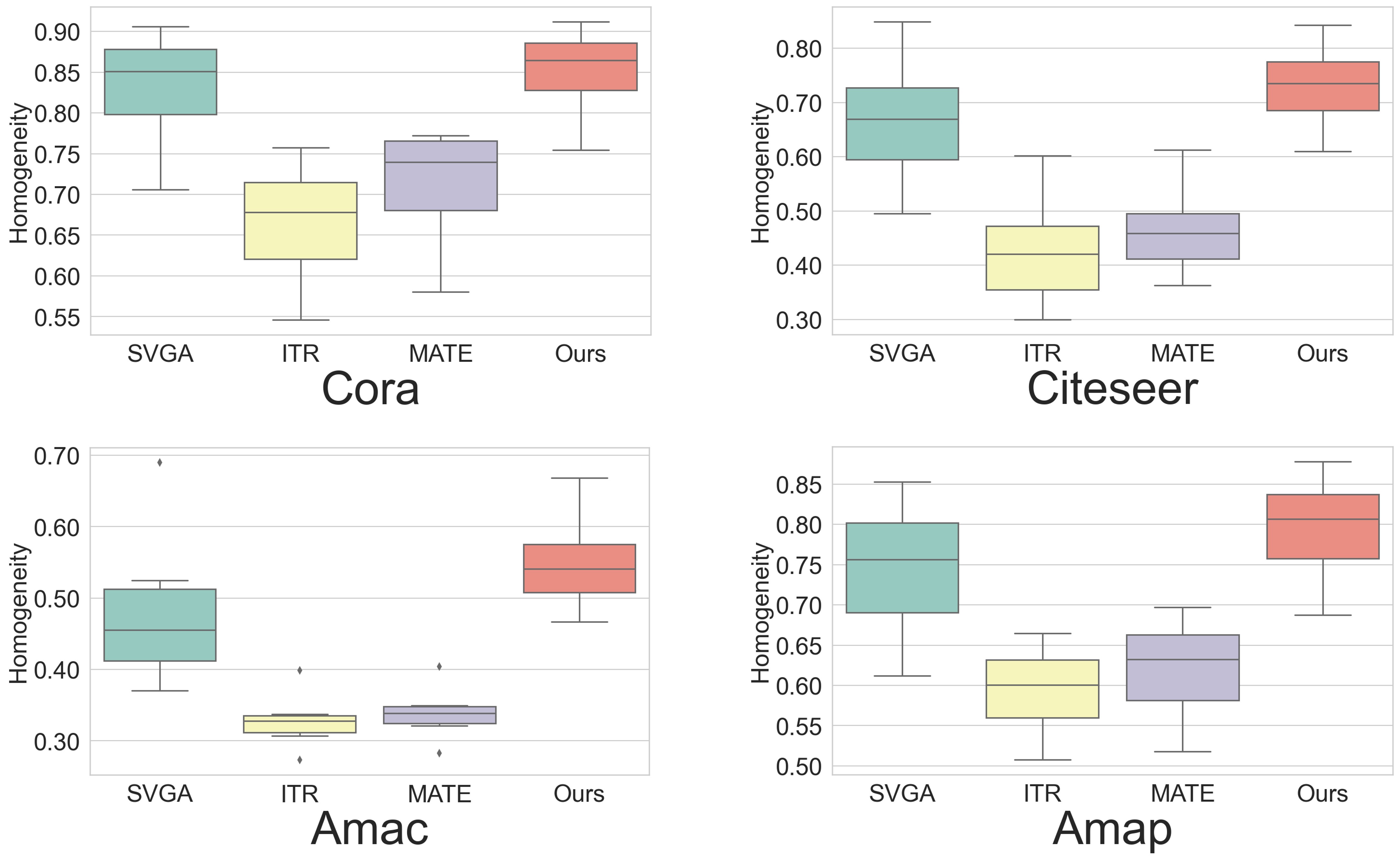}
\caption{Comparison of homogeneity of KNN graphs constructed by features reconstructed by four methods.}
\label{fig5}
\end{figure}

\subsection{Robustness under Varying Missing Rates (\textbf{RQ3})}

In Figure \ref{fig2}, we conducted a comprehensive analysis to evaluate the feature reconstruction performance of four main methods: SVGA, ITR, MATE, and TDAR across four datasets at varying missing data rates (0.2-0.8). The results clearly demonstrate that TDAR consistently achieves the best performance under different missing rates on the Recall@k and nDGC@k metrics. Further observations show a general trend of performance decline across all methods as the missing data rate increases. However, TDAR consistently maintains its superior performance regardless of the missing data rate. While MATE shows competitive performance with TDAR at very high missing rates (0.7-0.8), it significantly underperforms TDAR at other missing rates. In contrast, SVGA and ITR generally exhibit suboptimal performance, with their performance decline becoming more pronounced in scenarios of high missing data rates compared to TDAR.

\begin{figure*}[t]
\centering
\subfloat[Recall@10 on Cora.]{
\includegraphics[width=0.22\textwidth]{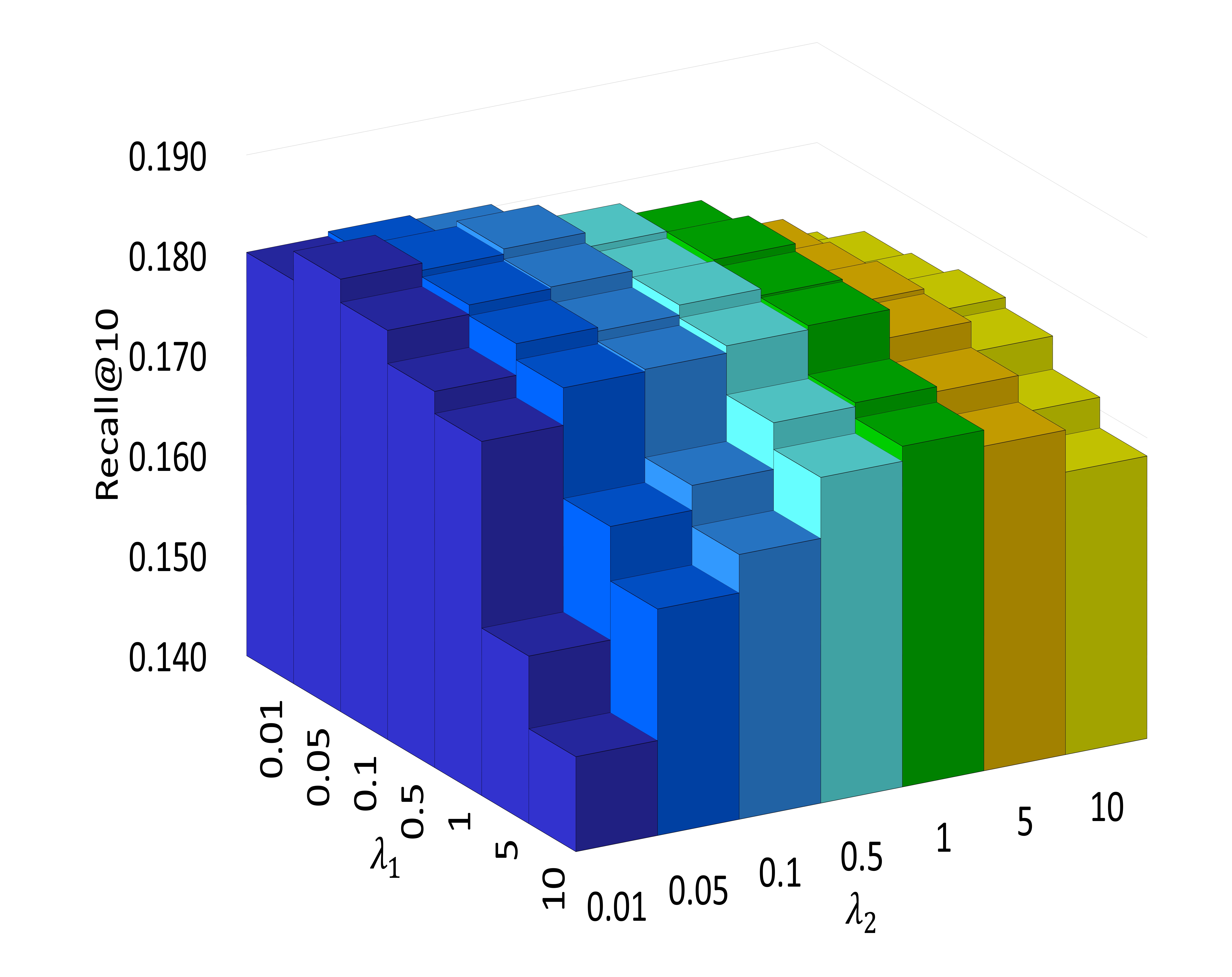}
}
\subfloat[ACC on Cora.]{
\includegraphics[width=0.22\textwidth]{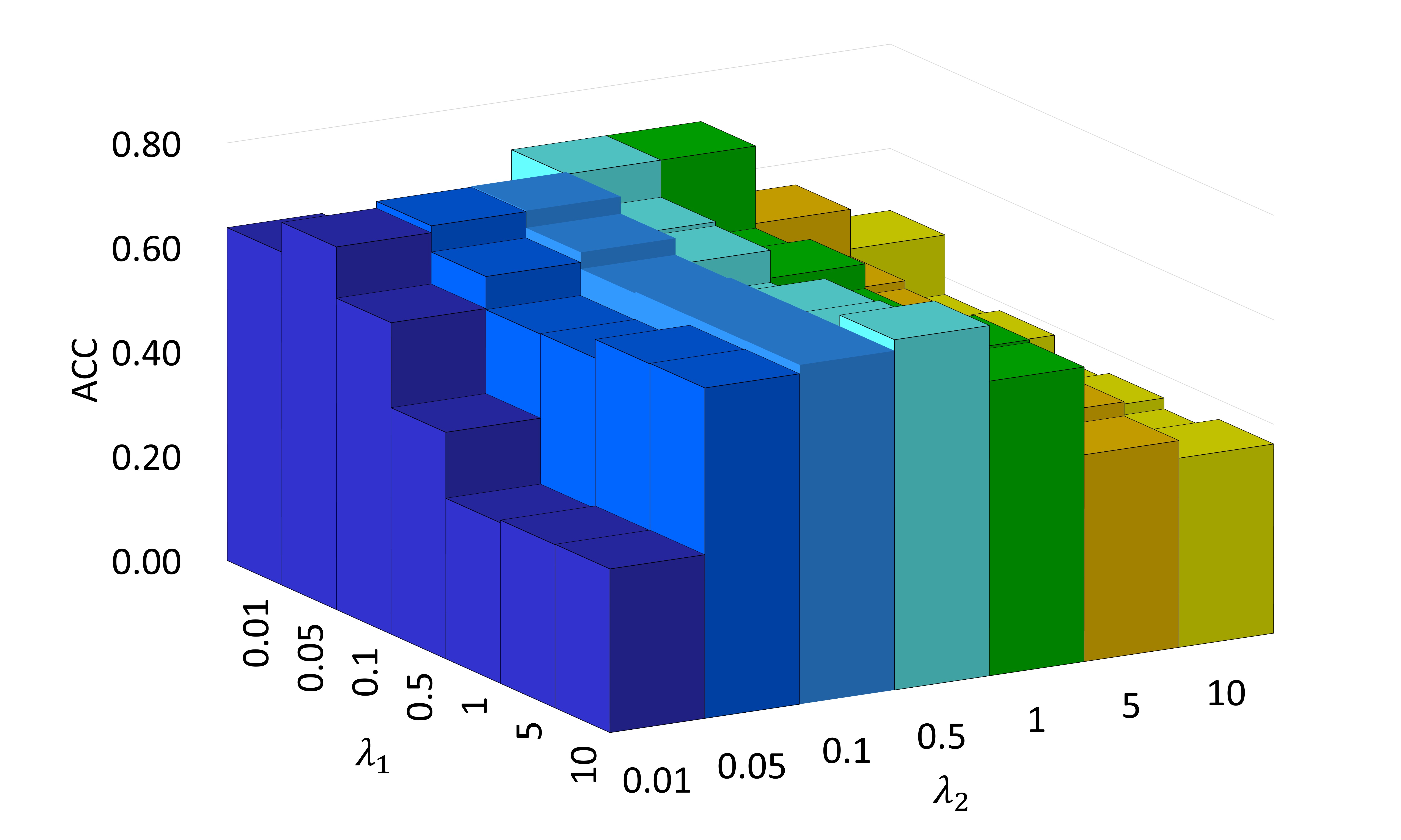}
}
\subfloat[Recall@10 on Citeser.]{
\includegraphics[width=0.22\textwidth]{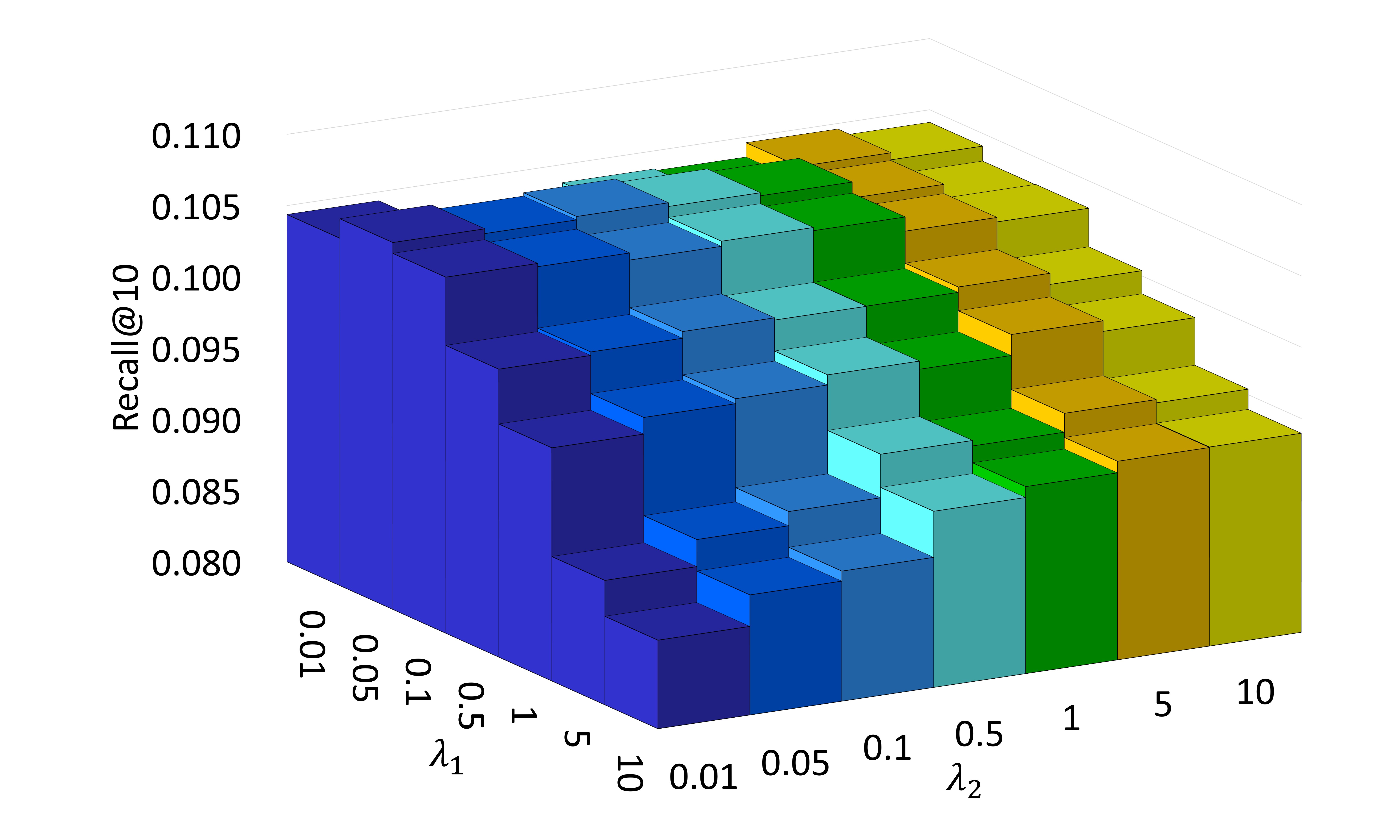}
}
\subfloat[ACC on Citeseer.]{
\includegraphics[width=0.22\textwidth]{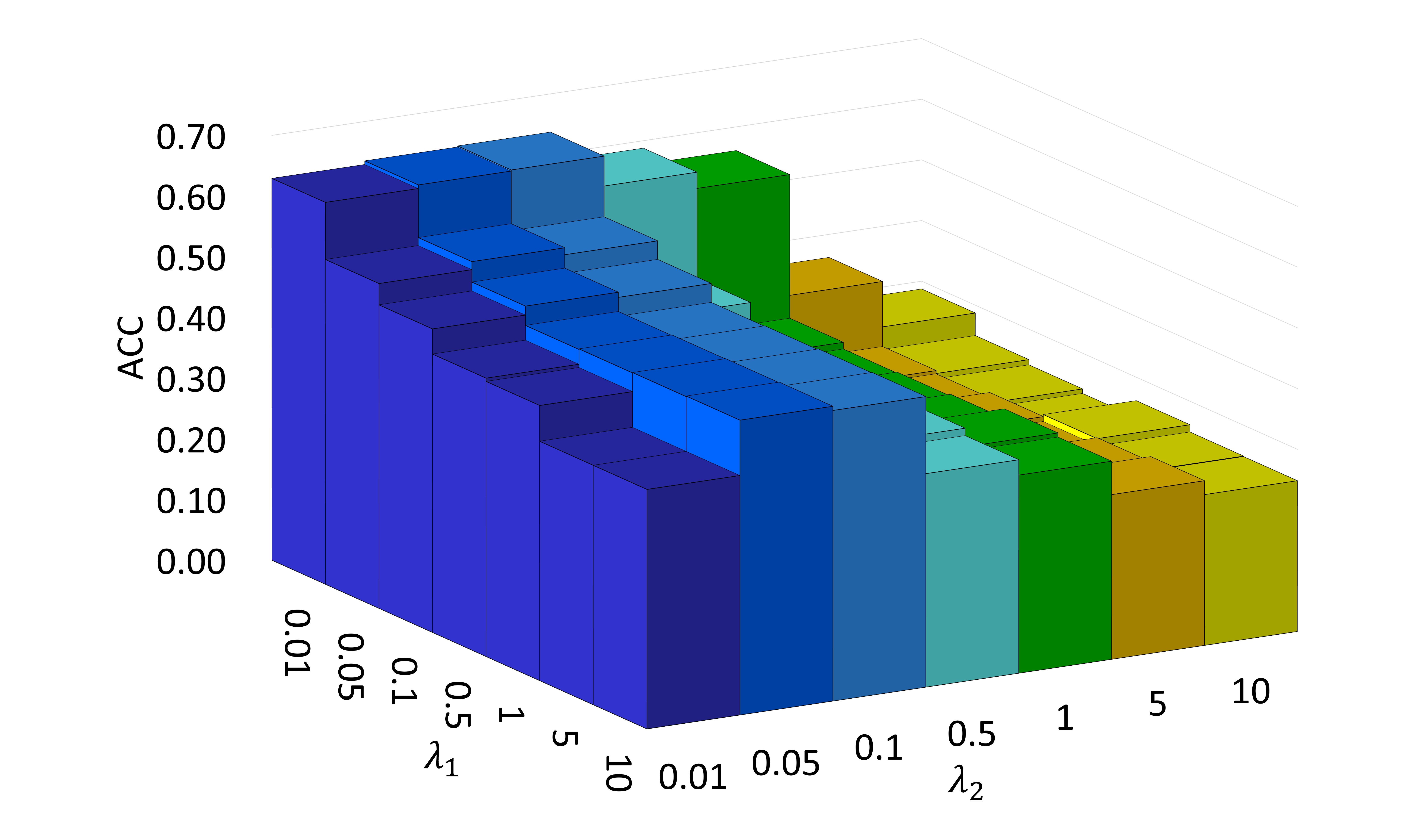}
}

\subfloat[Recall@10 on Amac.]{
\includegraphics[width=0.22\textwidth]{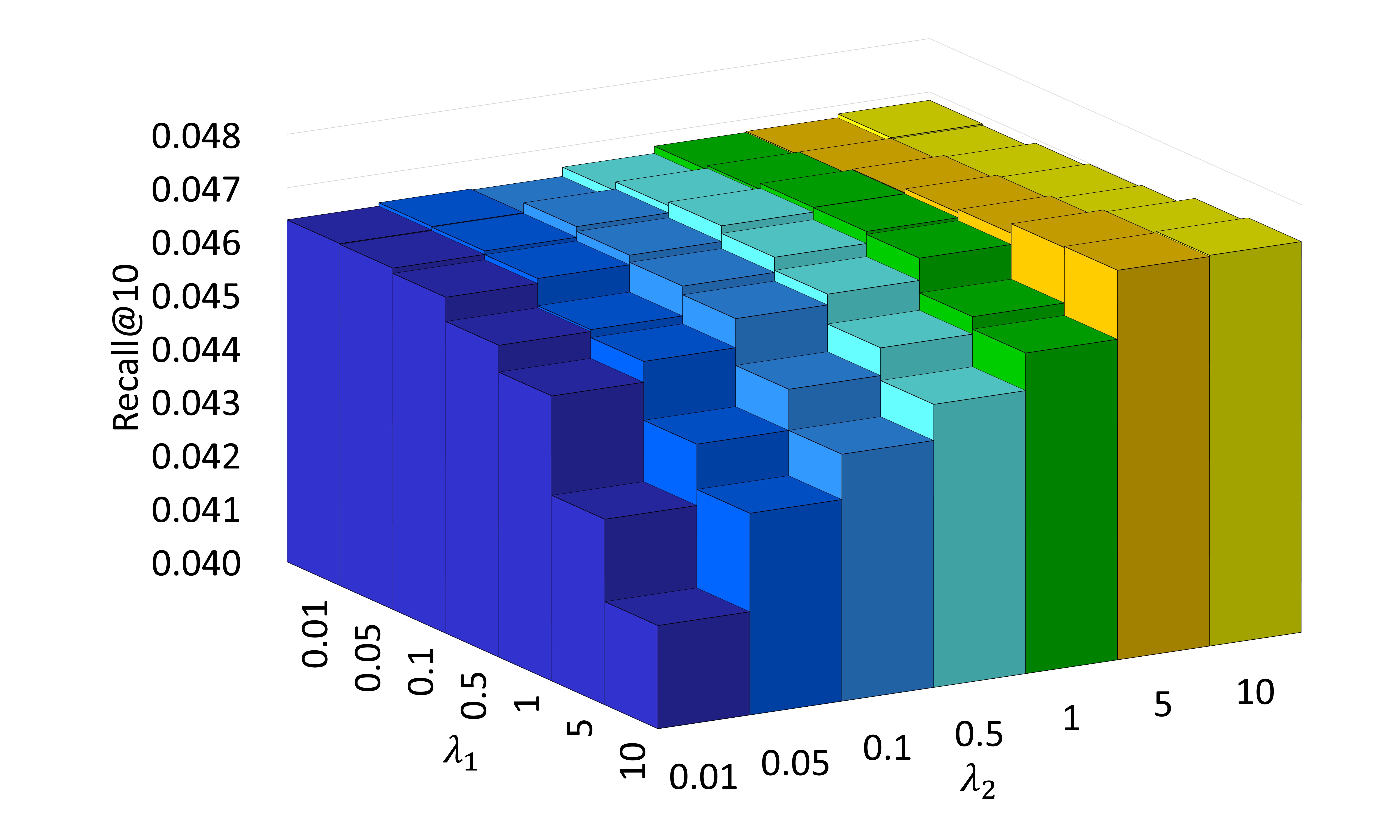}
}
\subfloat[ACC on Amac.]{
\includegraphics[width=0.22\textwidth]{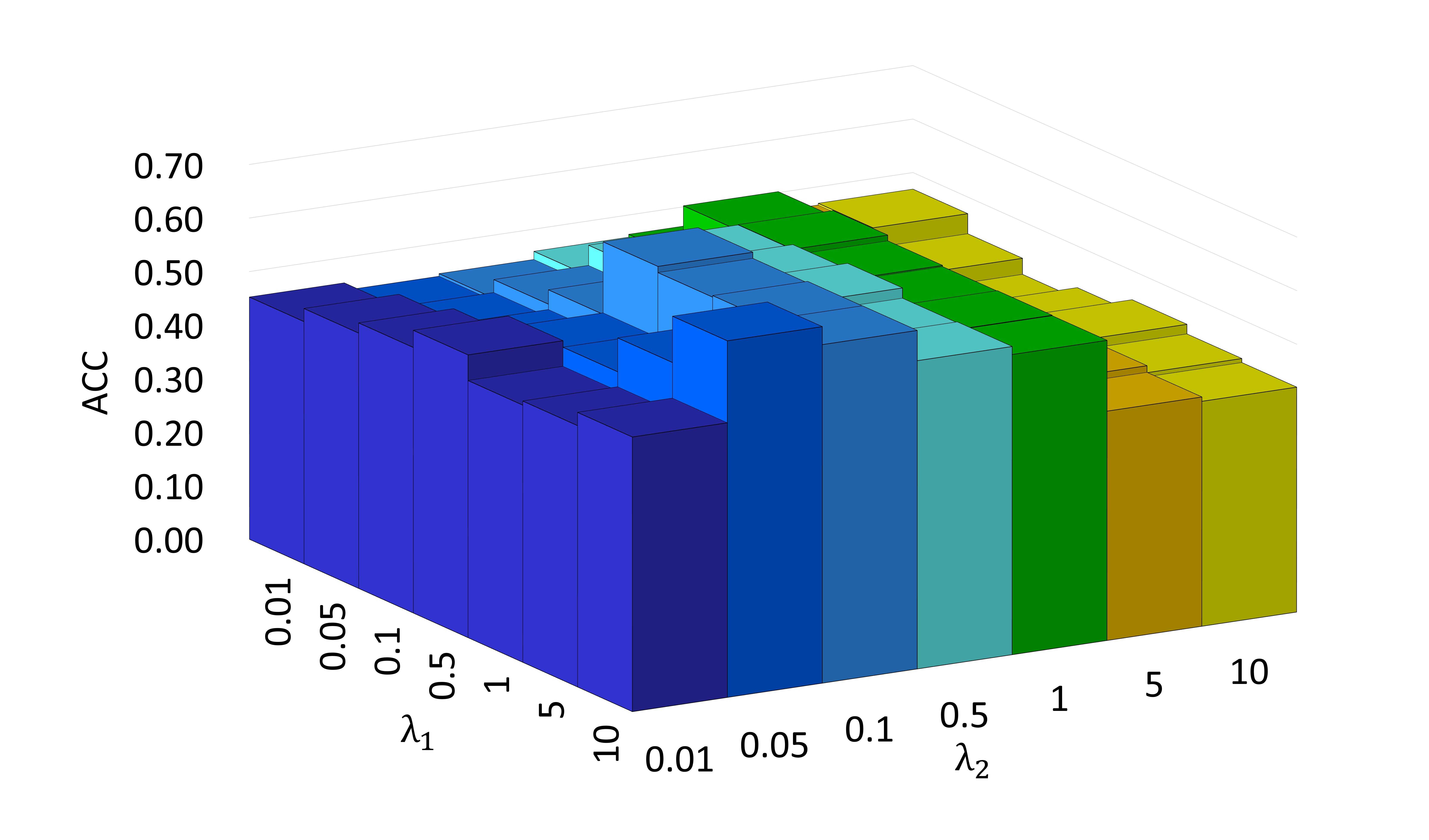}
}
\subfloat[Recall@10 on Amap.]{
\includegraphics[width=0.22\textwidth]{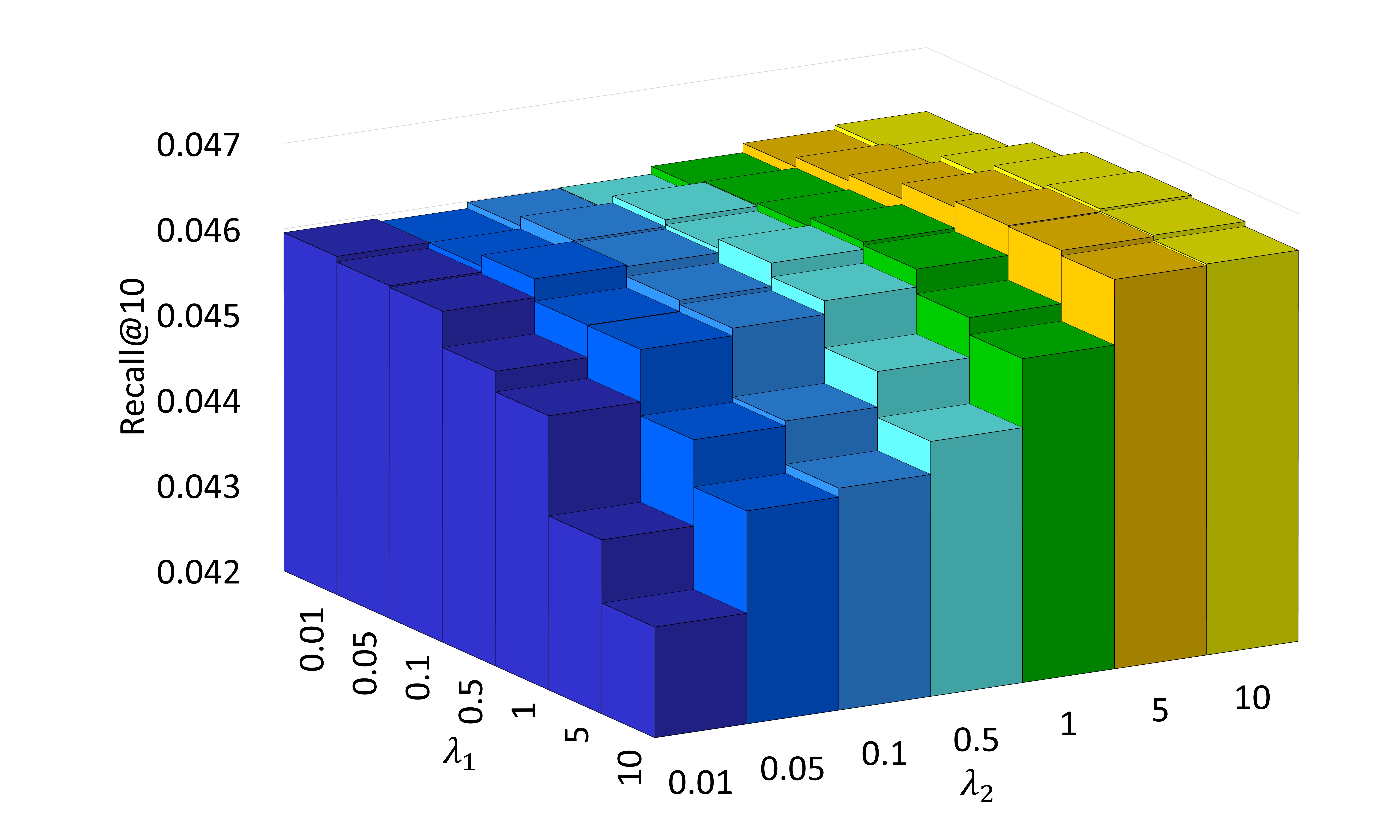}
}
\subfloat[ACC on Amap.]{
\includegraphics[width=0.22\textwidth]{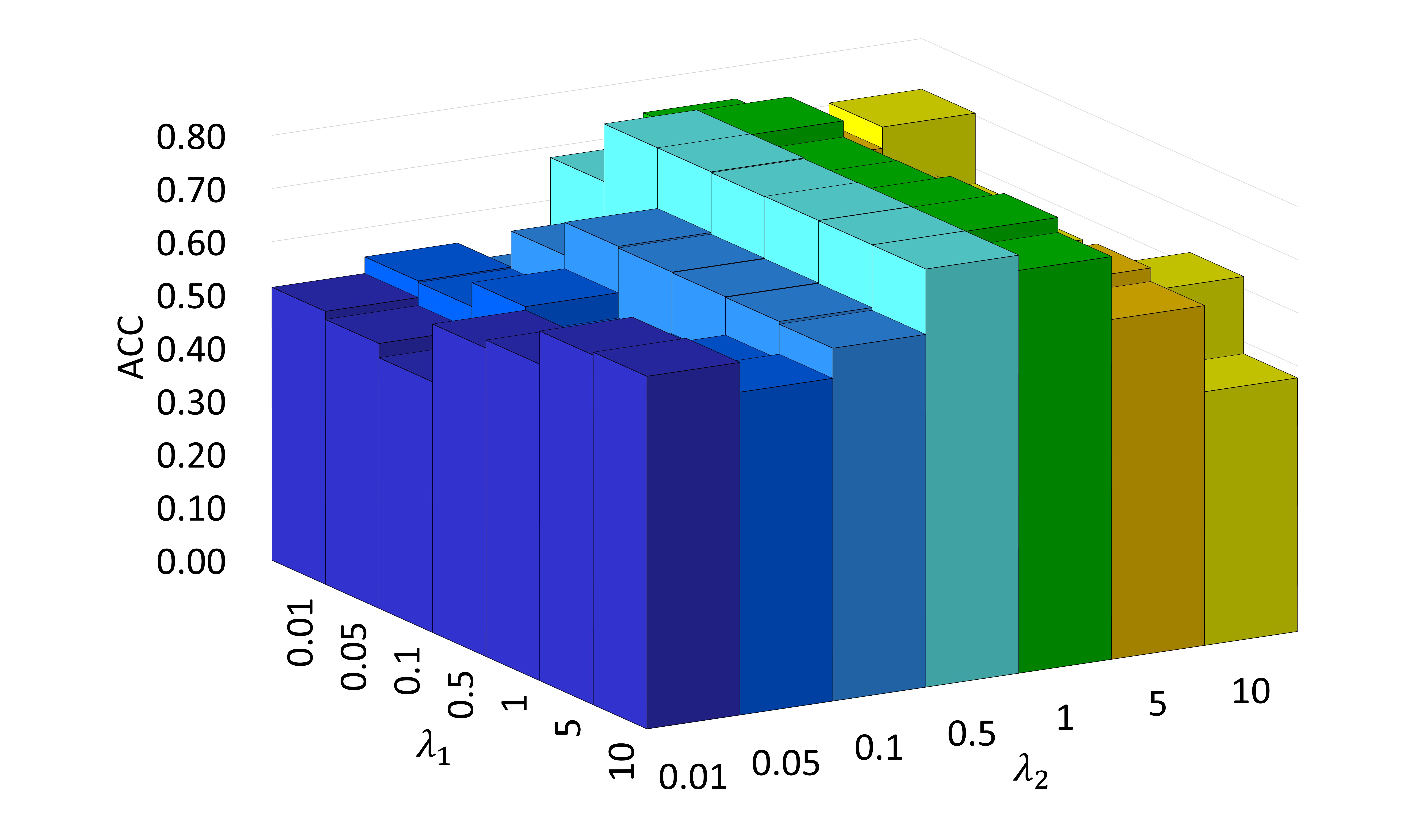}
}
\caption{Hyperparameter verification for attribute reconstruction using Recall@10 and node clustering using ACC metrics evaluating $\lambda_1$ and $\lambda_2$.}
\label{fig3}
\end{figure*}

\subsection{Homogeneity and Generalization Verification (\textbf{RQ4})}
High homogeneity indicates that the model excels in preserving the intrinsic class structure of the data, which is crucial for applications such as social network analysis where class structure often provides key insights for evaluation. Therefore, quantifying the homogeneity of reconstructed features will better understand the model's robustness and reliability. We analyze each dataset using a K-nearest neighbors (KNN) graph \cite{guo2003knn} derived from the reconstructed features, considering a range of neighborhood sizes specified by $k$ values from \{1, 10, 20, 50, 100, 200\}. The larger the neighborhood, the greater the impact on homogeneity. We use the homogeneity metric $\frac{\left|\left\{(v, w):(v, w) \in E \wedge y_v=y_w\right\}\right|}{|E|}$ \cite{zhu2020beyond} to evaluate the proportion of node pairs with the same label $y$ among all connected pairs in the graph. As shown in Figure \ref{fig5}, the $k$ value increases from the upper boundary to the lower boundary. The results show that our method exhibits excellent homogeneity on all datasets and significantly outperforms other compared methods.

Specifically, most comparison methods show a significant drop in homogeneity at larger $k$ values, but TDAR consistently maintains high homogeneity levels across all different $k$ values. This result highlights the excellent robustness of our TDAR, demonstrating its advantage in preserving the intrinsic class structure of the data.

\begin{figure}[t]
\centering
\subfloat[$l$ on Cora.]{
\includegraphics[width=0.22\textwidth]{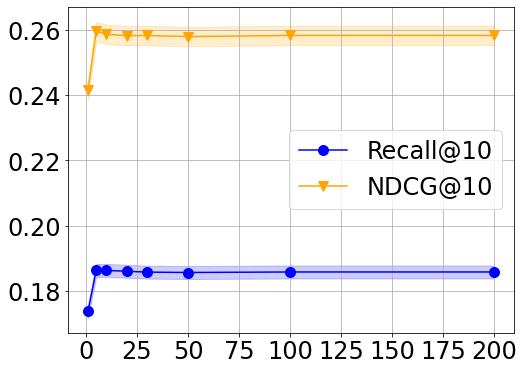}
}
\subfloat[$l$ on Citeseer.]{
\includegraphics[width=0.22\textwidth]{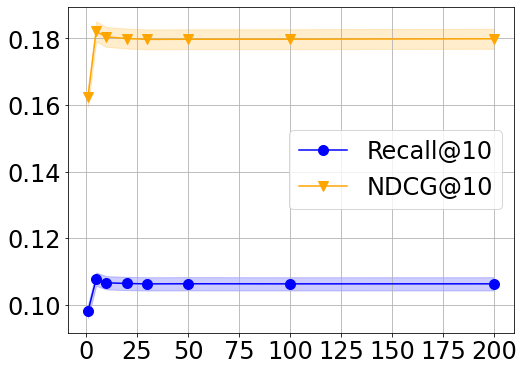}
}

\subfloat[$\epsilon$ on Cora.]{
\includegraphics[width=0.22\textwidth]{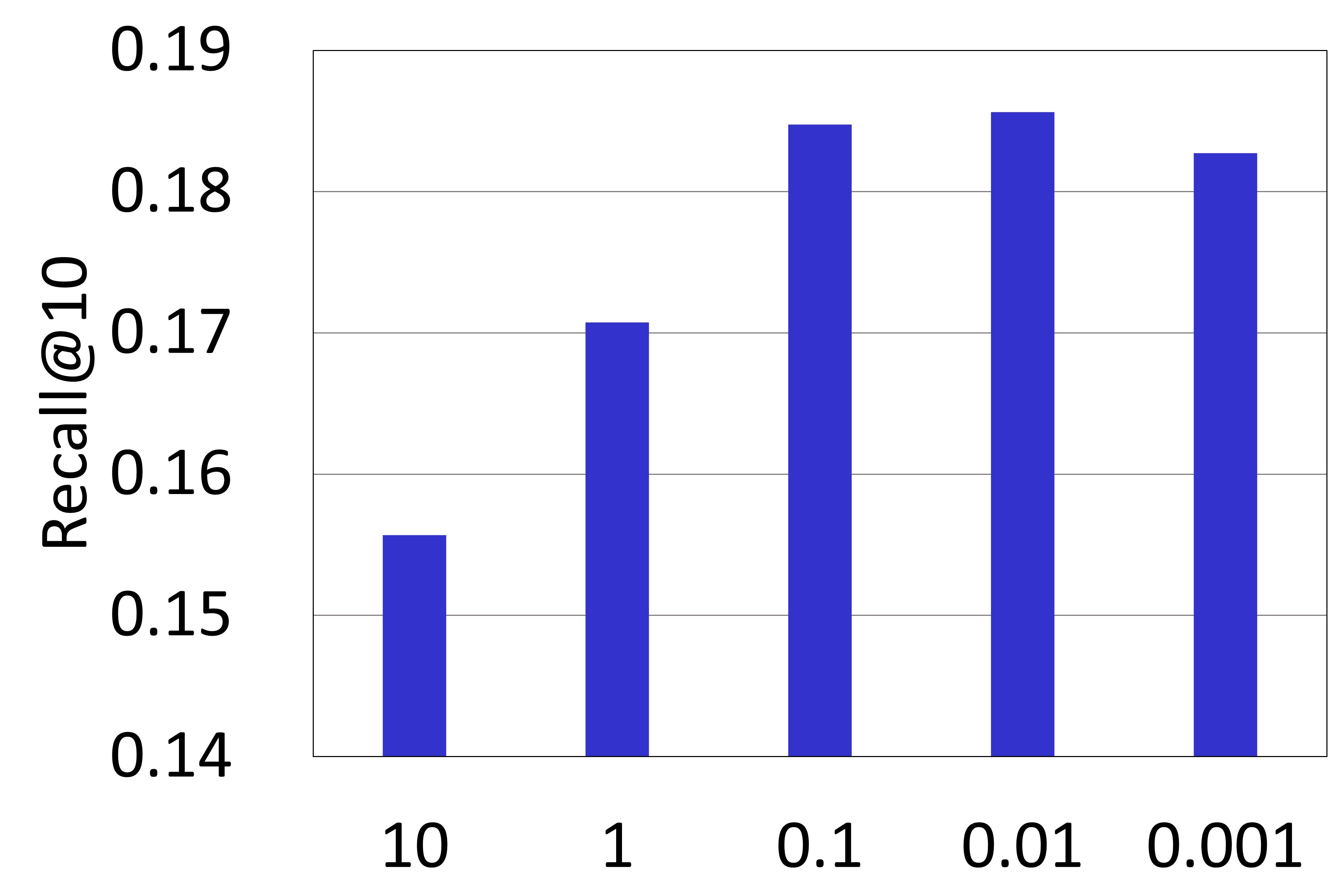}
}
\subfloat[$\epsilon$ on Citeseer.]{
\includegraphics[width=0.22\textwidth]{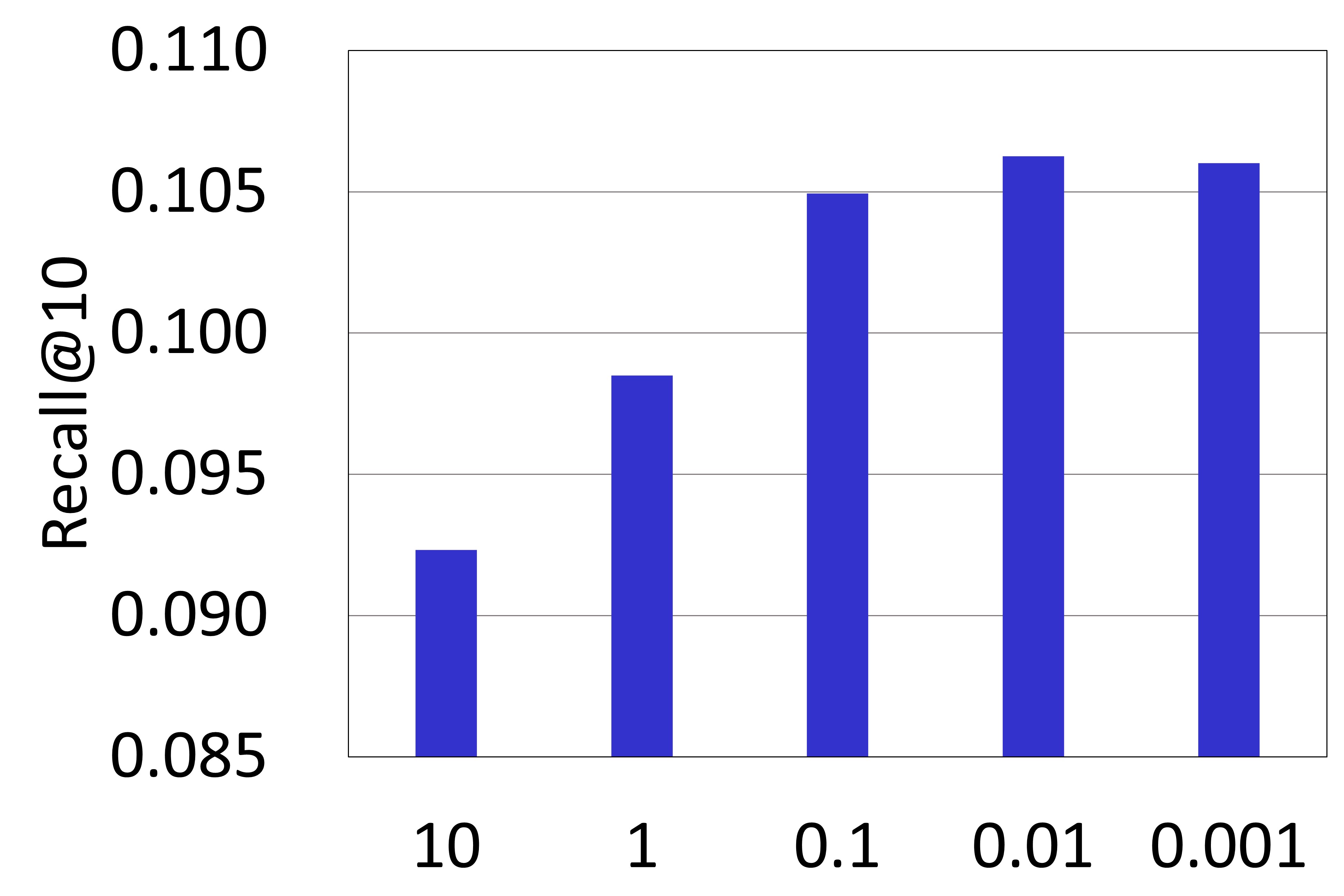}
}

\subfloat[$\alpha$ on Cora.]{
\includegraphics[width=0.22\textwidth]{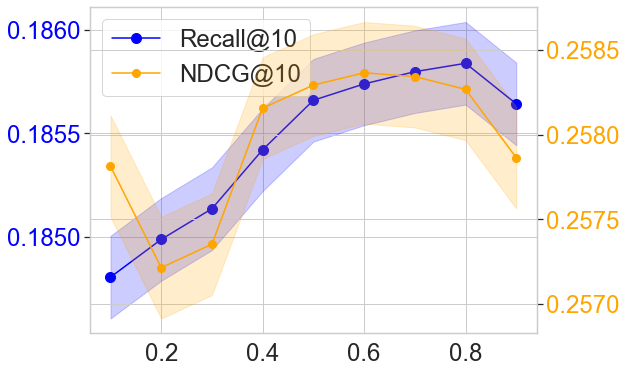}
}
\subfloat[$\alpha$ on Citeseer.]{
\includegraphics[width=0.22\textwidth]{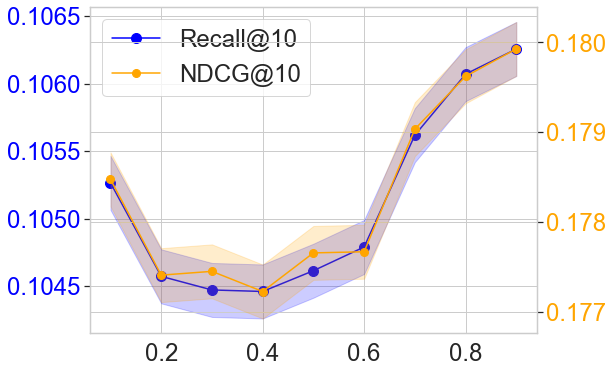}
}
\caption{Validation of remaining key hyperparameters $l$, $\epsilon$ and $\alpha$.}
\label{fig4}
\end{figure}

\subsection{Impact of Hyperparameters (\textbf{RQ5})}
The hyperparameters $\lambda_1$ and $\lambda_2$ play a pivotal role in harmonizing the overall optimization of TDAR. Figure \ref{fig3} provides an in-depth analysis to determine the optimal hyperparameter settings for $\lambda_1$ and $\lambda_2$. It exhibits two critical metrics: Recall@10 for feature reconstruction and AC for clustering. The analysis allows for the identification of the best hyperparameter values tailored to different datasets. Specifically, for Cora and Citeseer, the optimal setting for $\lambda_1$ is 0.1 and for $\lambda_2$ it is 0.05; for Amac, both parameters are optimally set at 10; for Amap, the ideal settings are 1 for $\lambda_1$ and 10 for $\lambda_2$. In node clustering, slight adjustments are recommended to achieve optimal performance: Cora keeps $\lambda_1$ at 0.05 while increasing $\lambda_2$ to 0.1; Citeseer reduces $\lambda_1$ to 0.01 while $\lambda_2$ remains at 0.1; Amac maintains $\lambda_1$ at 10 but decreases $\lambda_2$ to 0.1; Amap adjusts $\lambda_1$ to 0.05 and $\lambda_2$ to 0.5. These findings suggest that Cora and Citeseer respond well to smaller adjustments in the hyperparameter values, whereas Amac and Amap show a broader range of responses.

Furthermore, Figure \ref{fig4} delves into the complexities of other key hyperparameters, namely $l$, which controls the number of TAAP iterations, $\epsilon$, which optimizes the  latent embeddings for ESPC, and $\alpha$, the decay factor. The experiment focuses on the Cora and Citeseer datasets. The findings suggest that the hyperparameter $l$ performs best within the range of 5 to 10, as excessive smoothing may overlook outliers. $\epsilon$ yields optimal results at 0.01, and the ideal settings for $\alpha$ are pinpointed at 0.8 and 0.9.

\subsection{Visualization (\textbf{RQ6})}

\begin{figure}[t]
\centering
\includegraphics[width=0.45\textwidth]{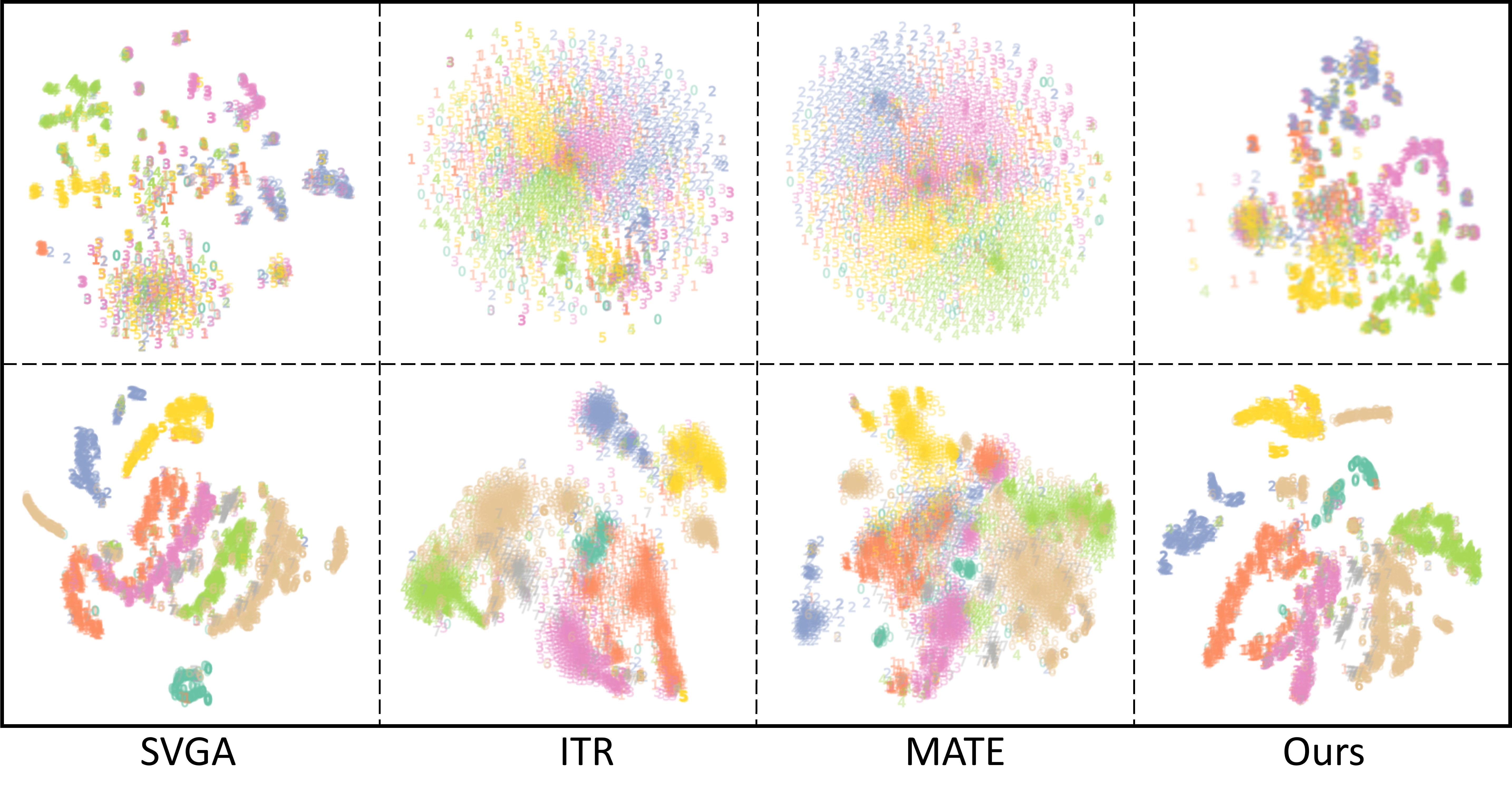}
\caption{Reconstructed feature dimensionality reduction visualization for Citeseer (1st Row) and Amap (2nd Row).}
\label{fig6}
\end{figure}

We showcased visualizations of attribute features generated by SVGA, ITR, MATE, and our TDAR, following t-SNE \cite{van2008visualizing} dimensionality reduction, applied to the Citeseer and Amac datasets. Figure \ref{fig6} showed SVGA's data points clustering with unclear boundaries and some overlap, while ITR and MATE's points were more diffused with less distinct communities.  Contrarily, TDAR displayed pronounced clustering with distinct community boundaries, highlighting a superior level of distinctiveness among various groups.

\begin{figure}[t]
\centering
\includegraphics[width=0.45\textwidth]{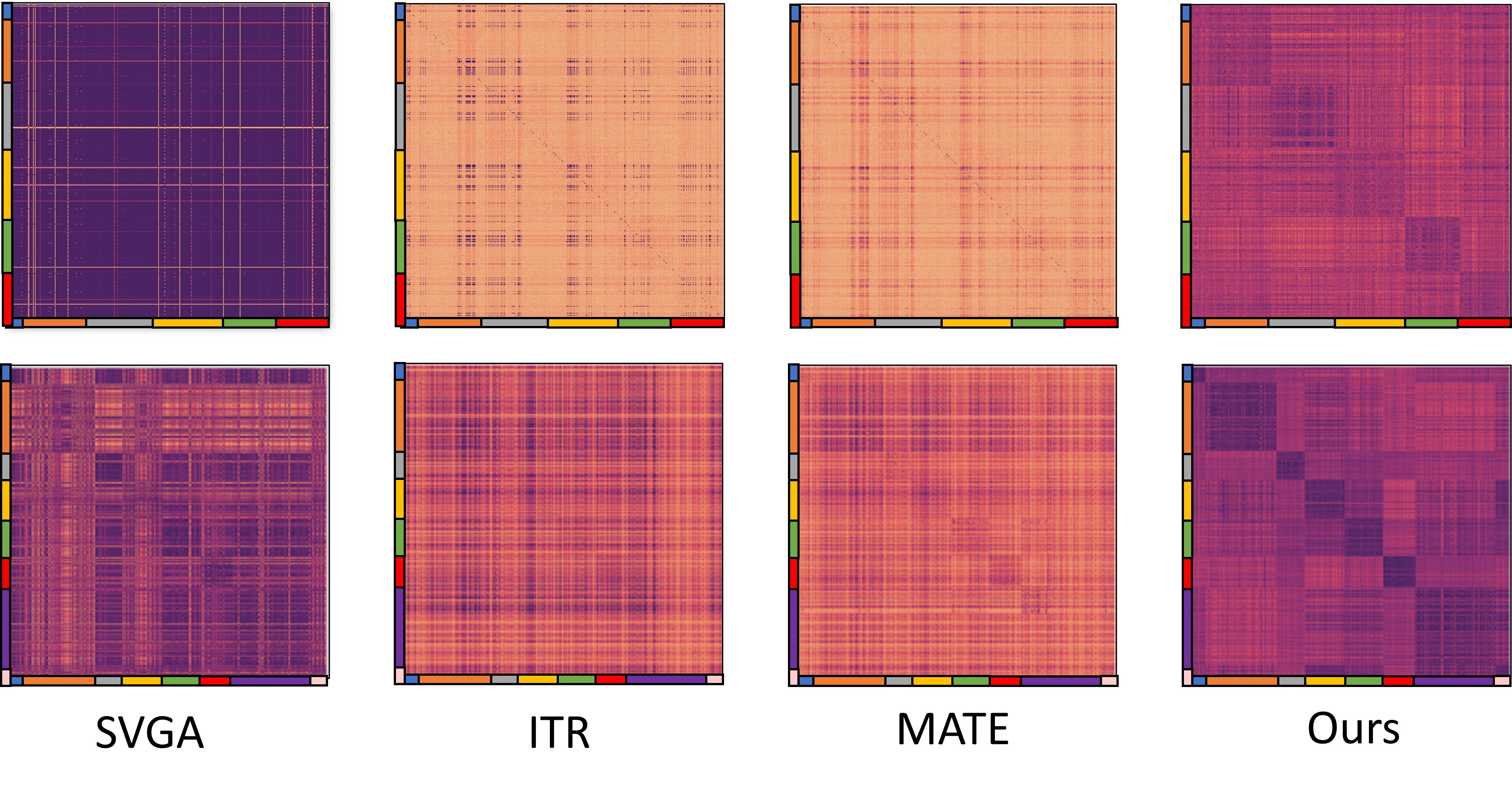}
\caption{Similarity heatmaps for Citeseer (1st Row) and Amap (2nd Row).}
\label{fig7}
\end{figure}

Figure \ref{fig7} provided a detailed comparison of the reconstructed feature similarity matrices generated by these methods on the Citeseer and Amac datasets. The heatmap clearly revealed varying structural patterns for each method. Specifically, SVGA, ITR, and MATE displayed some degree of structure, but with certain mismatched areas, suggesting that these methods might not effectively capture the intra-class information of the datasets during the feature generation process. On the other hand, our TDAR showed a more compact and contiguous structural pattern, particularly along the main diagonal, indicating strong clustering ability. This implies that our method better preserved the original structure of the data in the feature space, demonstrating robust noise resistance and excellent preservation of intra-class information.

\subsection{Case Analysis (\textbf{RQ7})}

\begin{figure}[t]
\centering
\includegraphics[width=0.45\textwidth]{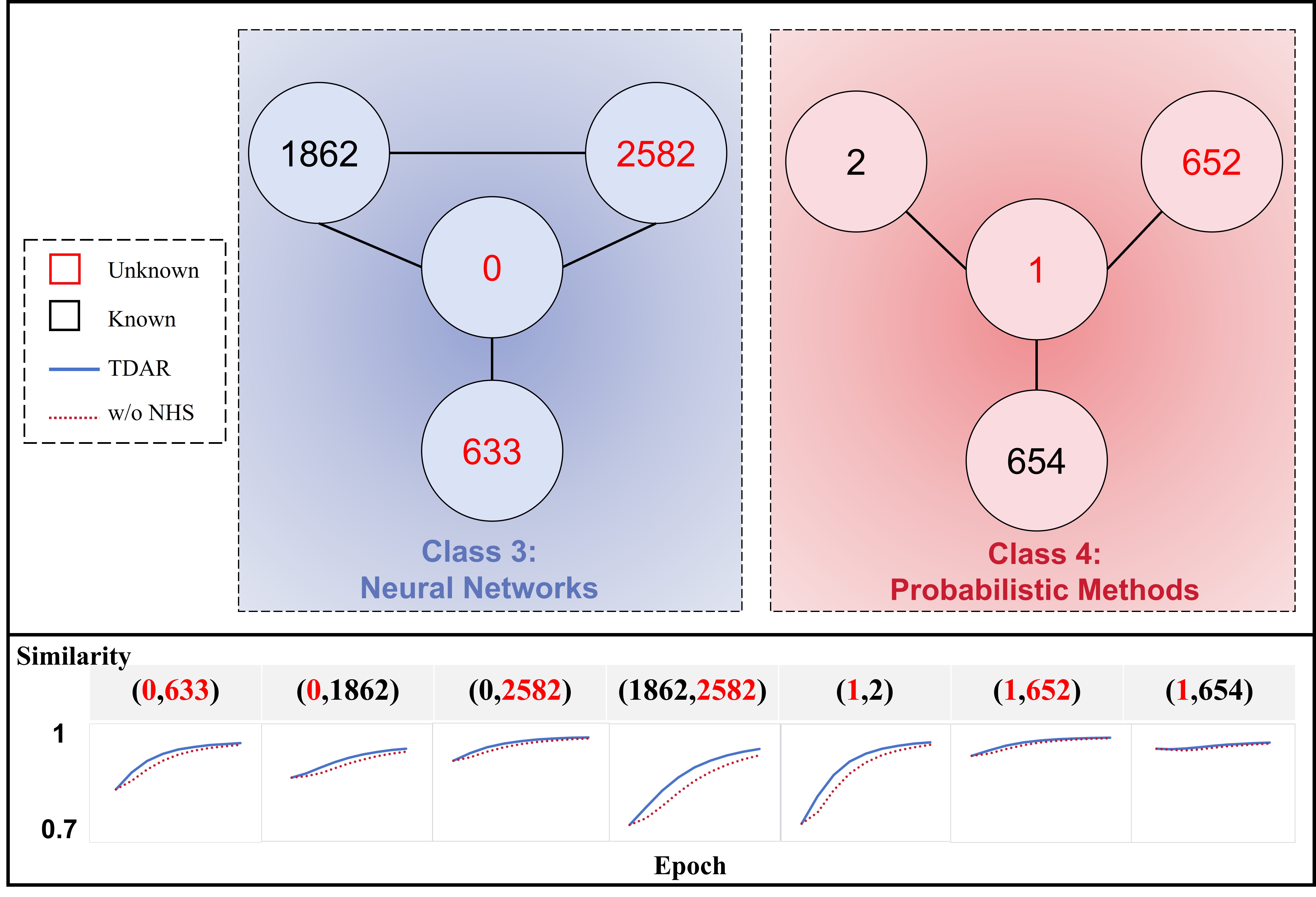}
\caption{A case of attribute feature similarity for selected nodes on Cora.}
\label{fig8}
\end{figure}

To deepen our understanding of TDAR's efficacy, particularly in the context of graph datasets with diverse component sizes, we extended our case study to encompass small components within the Cora dataset. This is vital for demonstrating the global propagation benefits posited in TDAR, especially in enhancing the performance of nodes in smaller graph components. In our initial analysis, as illustrated in Figure \ref{fig8}, we focused on TDAR's effect on node feature similarity during the training process. This involved examining nodes with both known (2, 654, 1862) and unknown (0, 1, 633, 652, 2582) attributes, specifically noting the absence of direct edges between nodes of classes 3 and 4. We observed a general increase in feature similarity as training progressed, indicating that TDAR enhances similarity in latent features among connected nodes across iterations. Additionally, the removal of the NHS strategy resulted in a significant decrease in similarity, underscoring the importance of NHS in fostering node homogeneity in the embedding space and effectively clustering nodes, even those with missing attributes.

\section{Conclusion}

The increasing volume of interactive data from SIoT devices typically forms complex graphs, where missing node attributes pose significant challenges to analysis and prediction tasks. To address these issues, we present the TDAR framework, which integrates several key strategies to improve attribute reconstruction and overall graph learning. First, TAAP leverages the graph’s structural connectivity to propagate and estimate missing attributes, refining initial feature representations. Next, ESPC dynamically balances the embedding space by adjusting attention weights, ensuring that both global and local structural dependencies are captured effectively. Additionally, NHS and NLSC ensure that similar nodes are correctly positioned in the embedding space while mitigating the influence of non-connected yet similar node pairs, thereby preserving the graph's structural integrity. Our extensive experiments across multiple datasets demonstrate that TDAR significantly outperforms current benchmarks in attribute reconstruction, node classification, and clustering tasks.

Looking ahead, future research avenues are promising and manifold. A key direction is the exploration of TDAR's scalability and effectiveness across larger and more complex graph datasets. This would not only extend the framework's applicability but also provide insights into its performance in more demanding scenarios. Additionally, there is significant potential for adapting TDAR to accommodate various graph types, including but not limited to, heterogeneous and dynamic graphs. Such expansions could enhance the framework's utility in addressing the nuances of graph diversity and temporal changes. Moreover, refining TDAR to tackle the challenges posed by heterogeneity and distributional bias in graphs presents an exciting frontier for further research.

\bibliographystyle{IEEEtran}
\bibliography{bib.bib}

\begin{IEEEbiography}[{\includegraphics[width=1in,height=1.25in,clip,keepaspectratio]{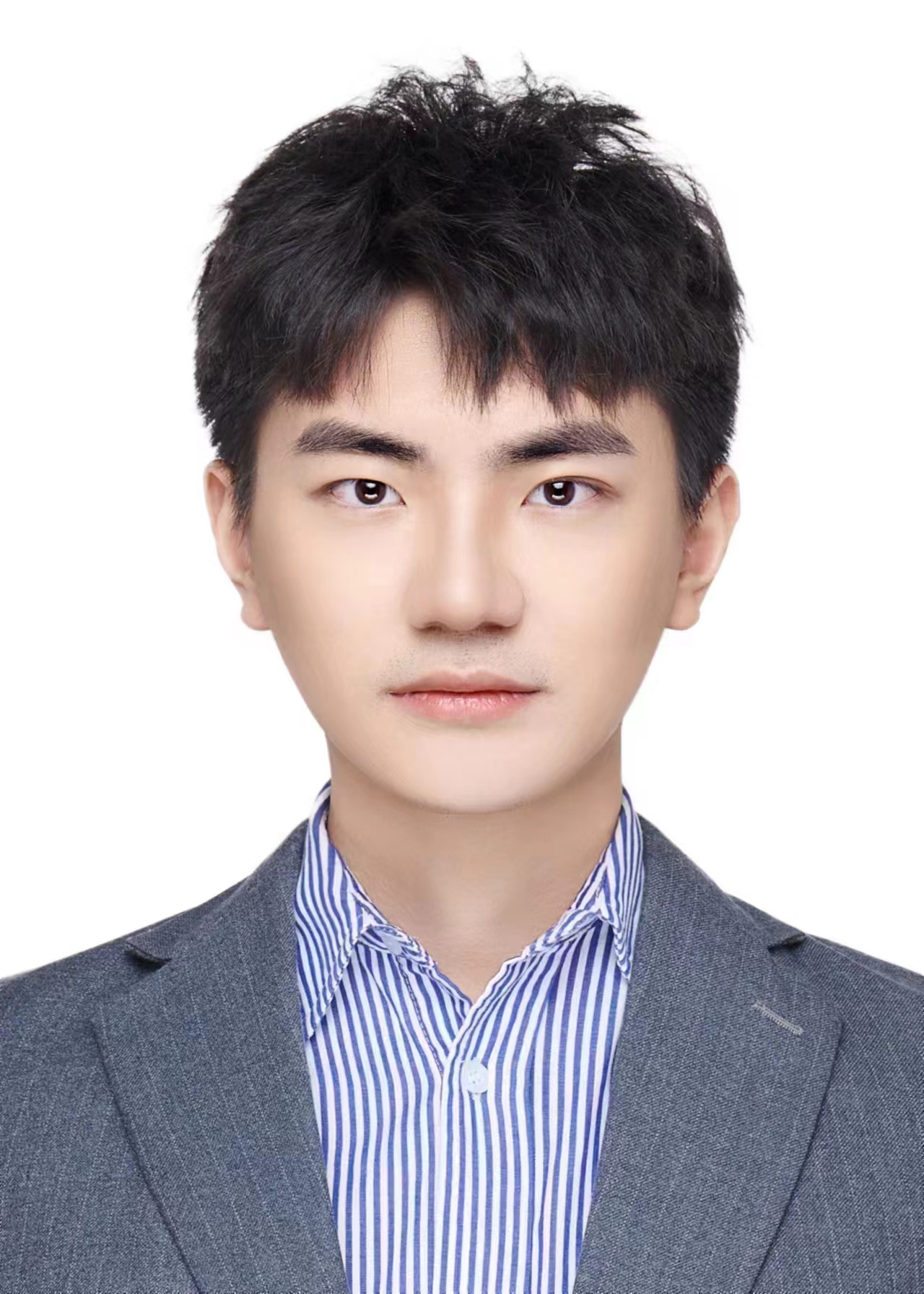}}]
{Mengran Li} is currently pursuing a Ph.D. degree at the Guangdong Key Laboratory of Intelligent Transportation Systems, School of Intelligent Systems Engineering, Sun Yat-sen University, Shenzhen, P.R. China. He received an M.S. degree in Control Science and Engineering from the Beijing Key Laboratory of Multimedia and Intelligent Software Technology, Beijing University of Technology, Beijing, P.R. China, in 2023. His research interests include graph neural networks, data mining, and complex network optimization.
\end{IEEEbiography}

\begin{IEEEbiography}[{\includegraphics[width=1in,height=1.25in,clip,keepaspectratio]{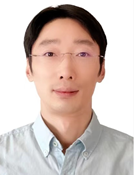}}]
{Junzhou Chen} received his Ph.D. in Computer Science and Engineering from the Chinese University of Hong Kong in 2008. Between March 2009 and February 2019, he served as a Lecturer and later as an Associate Professor at the School of Information Science and Technology at Southwest Jiaotong University. He is currently an Associate Professor at School of Intelligent Systems Engineering at Shenzhen Campus of Sun Yat-sen University. His research interests include computer vision, machine learning and intelligent transportation systems.
\end{IEEEbiography}

\begin{IEEEbiography}[{\includegraphics[width=1in,height=1.25in,clip,keepaspectratio]{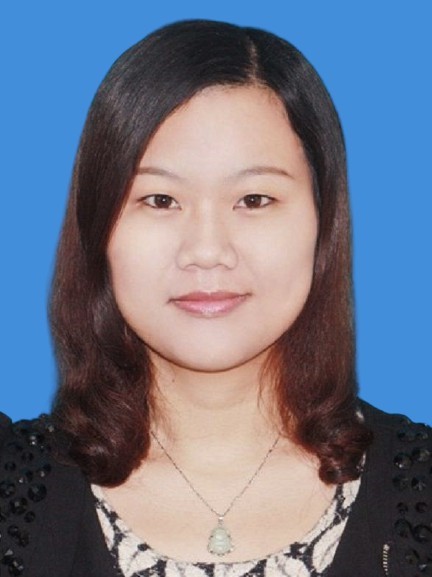}}]
{Chenyun Yu} received the BE and ME degrees in computer science from Chongqing University, Chongqing, China, in 2011 and 2014, respectively, and the PhD degree in computer science from the City University of Hong Kong, in 2018.
She is currently an Assistant Professor at School of Intelligent Systems Engineering at Shenzhen Campus of Sun Yat-sen University. Her research interests include data processing, query optimization, and largescale machine learning.
\end{IEEEbiography}

\begin{IEEEbiography}[{\includegraphics[width=1in,height=1.7in,clip,keepaspectratio]{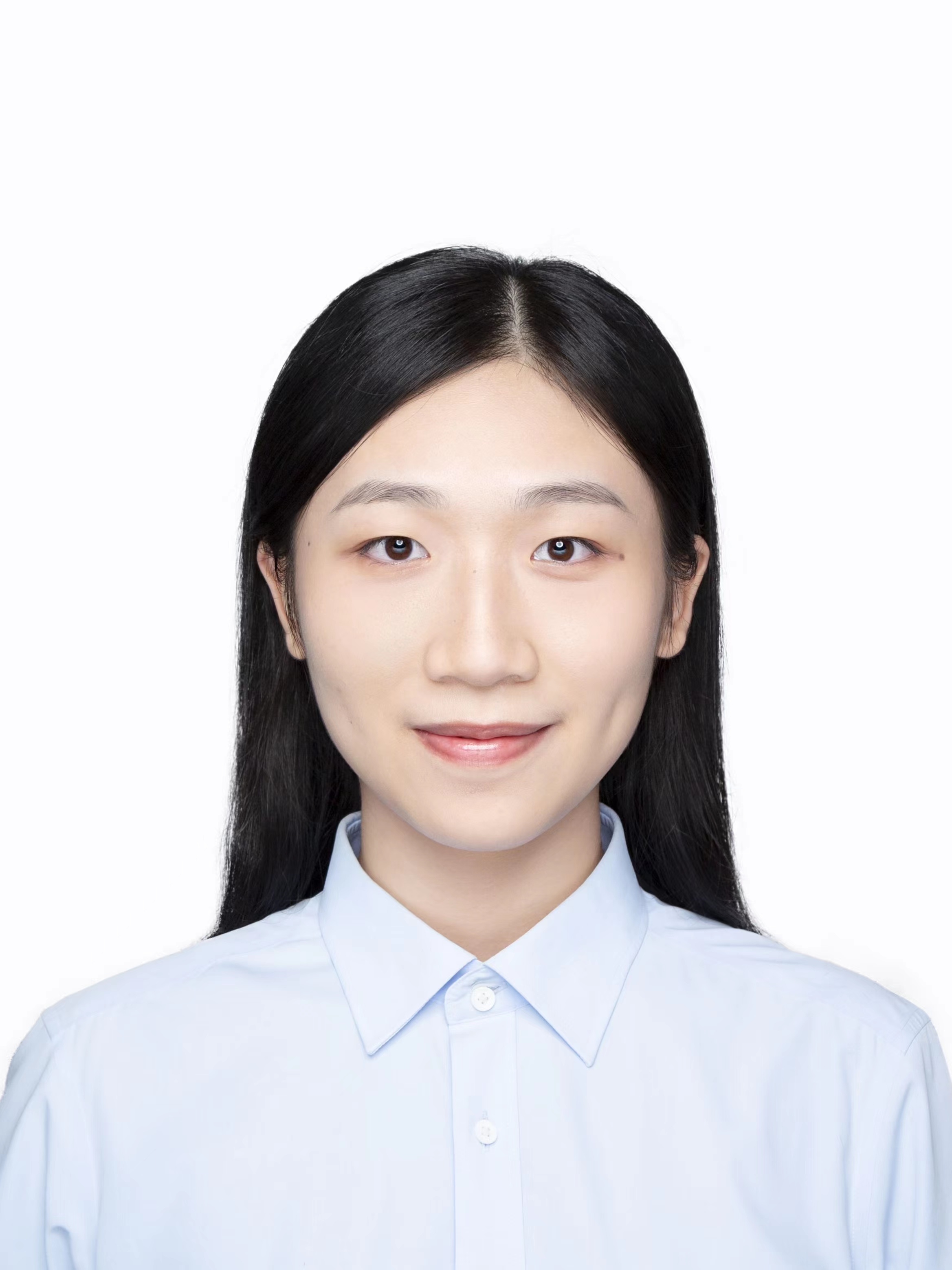}}]
{Guanying Jiang} received the B.S. and M.Ec. degrees in Statistics from Guangzhou University and Jinan University, Guangzhou, China, in 2017 and 2020, respectively. She has also been a Visiting Research Assistant at the School of Intelligent Systems Engineering at Sun Yat-sen University, Guangzhou, China. Since 2020, she has been a Senior Algorithm R\&D Engineer at Baidu Inc. Her research interests include artificial intelligence, intelligent transportation systems, and urban mobility modeling.
\end{IEEEbiography}

\begin{IEEEbiography}[{\includegraphics[width=1in,height=1.5in,clip,keepaspectratio]{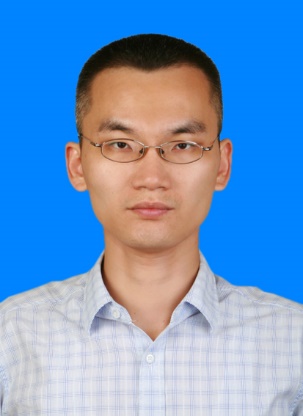}}]
{Ronghui Zhang} received a B.Sc. (Eng.) from the Department of Automation Science and Electrical Engineering, Hebei University, Baoding, P.R. China, in 2003, an M.S. degree in Vehicle Application Engineering from Jilin University, Changchun, P.R. China, in 2006, and a Ph.D. (Eng.) in Mechanical \& Electrical Engineering from Changchun Institute of Optics, Fine Mechanics and Physics, the Chinese Academy of Sciences, Changchun, P.R. China, in 2009. After finishing his post-doctoral research work at INRIA, Paris, France, in February 2011, he is currently an Associate Professor with Guangdong Key Laboratory of Intelligent Transportation System, School of Intelligent Systems Engineering, Shenzhen Campus of Sun Yat-sen University, Shenzhen, Guangdong, P.R. China. His current research interests include computer vision, intelligent control, and ITS. 
\end{IEEEbiography}

\begin{IEEEbiography}[{\includegraphics[width=1in,height=1.5in,clip,keepaspectratio]{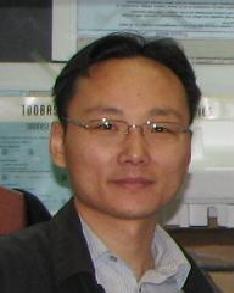}}]
{Yanming Shen} received his B.S. degree in automation from Tsinghua University in 2000 and his Ph.D. degree from the Department of Electrical and Computer Engineering at Polytechnic University (now NYU Tandon School of Engineering) in 2007. He is currently a Professor at the School of Computer Science and Technology, Dalian University of Technology, China. His research interests include big data analytics, distributed systems, and networking. Shen was the recipient of the 2011 Best Paper Award for Multimedia Communications, awarded by the IEEE Communications Society.
\end{IEEEbiography}

\begin{IEEEbiography}[{\includegraphics[width=1in,height=1.5in,clip,keepaspectratio]{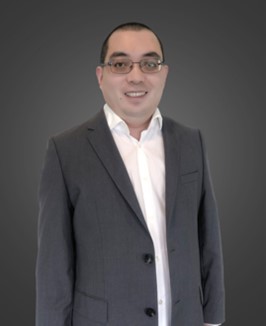}}]
{Houbing Herbert Song} (M’12–SM’14-F’23) received the Ph.D. degree in electrical engineering from the University of Virginia, Charlottesville, VA, in August 2012.
He is currently a Professor, the Founding Director of the NSF Center for Aviation Big Data Analytics (Planning), the Associate Director for Leadership of the DOT Transportation Cybersecurity Center for Advanced Research and Education (Tier 1 Center), and the Director of the Security and Optimization for Networked Globe Laboratory (SONG Lab, www.SONGLab.us), University of Maryland, Baltimore County (UMBC), Baltimore, MD. Prior to joining UMBC, he was a Tenured Associate Professor of Electrical Engineering and Computer Science at Embry-Riddle Aeronautical University, Daytona Beach, FL. 
Dr. Song is an IEEE Fellow (for contributions to big data analytics and integration of AI with Internet of Things), an Asia-Pacific Artificial Intelligence Association (AAIA) Fellow, an ACM Distinguished Member (for outstanding scientific contributions to computing), and a Full Member of Sigma Xi. Dr. Song has been a Highly Cited Researcher identified by Web of Science since 2021. He is an ACM Distinguished Speaker (2020-present), an IEEE Computer Society Distinguished Visitor (2024-present), an IEEE Communications Society (ComSoc) Distinguished Lecturer (2024-present), an IEEE Vehicular Technology Society (VTS) Distinguished Lecturer (2023-present) and an IEEE Systems Council Distinguished Lecturer (2023-present). 
His research interests include AI/machine learning/big data analytics, cyber-physical systems/internet of things, and cybersecurity and privacy. 
\end{IEEEbiography}

\vfill

\end{document}